%% file: neurips_2026.tex
\documentclass{article}
    \PassOptionsToPackage{numbers, compress}{natbib}

\usepackage[most]{tcolorbox}  %

\usepackage{amsmath}
\usepackage{amssymb}
\usepackage{mathtools}
\usepackage{amsthm}
\usepackage{nicefrac}
\usepackage{float}

\definecolor{dgreen}{rgb}{0.0, 0.52, 0.34}

\theoremstyle{plain}

\theoremstyle{definition}

\theoremstyle{remark}

\input{preamble/math_commands}

\usepackage{url}

\usepackage{booktabs}
\usepackage{graphicx}
\usepackage{xspace}

\usepackage{titletoc}
\contentsuse{appendix}{toc}

\titlecontents{section}
  [0em]{}{\contentslabel{2.0em}}{} {\hfill\contentspage}
\titlecontents{subsection}
  [2.0em]{}{\contentslabel{2.6em}}{} {\hfill\contentspage}

\usepackage{float}
\usepackage{multirow}
\usepackage{multicol}

\usepackage{pifont}
\usepackage{xcolor}
\usepackage{soul}
\usepackage{subcaption}
\usepackage{ragged2e}
\usepackage{fontawesome5} %

\newcommand{\blue}[1]{\textcolor{black}{#1}}
\usepackage{xcolor}
\newenvironment{blueblock}
  {\begingroup\color{black}}
  {\endgroup}

\usepackage{caption}

\usepackage{wrapfig}  %

\usepackage{booktabs,tabularx,xcolor,pifont,array}
\usepackage{arydshln}
\newcommand{\cmark}{\textcolor{green!60!black}{\ding{51}}}
\newcommand{\xmark}{\textcolor{red!70!black}{\ding{55}}}

\newcommand{\cat}{\textsc{cat}\xspace}
\newcommand{\cats}{\textsc{cat}s\xspace}

\newcommand{\Cats}{\textsc{Cat}s\xspace}

\newcommand{\ud}[1]{\underline{#1}}

\usepackage{listings}
\usepackage{xcolor}

\definecolor{codegray}{rgb}{0.5,0.5,0.5}
\definecolor{codepurple}{rgb}{0.58,0,0.82}
\definecolor{backcolour}{rgb}{0.95,0.95,0.92}

\lstdefinestyle{pytorch}{
    backgroundcolor=\color{backcolour},   
    commentstyle=\color{codegray}\ttfamily,
    keywordstyle=\color{blue}\ttfamily,
    numberstyle=\tiny\color{gray},
    stringstyle=\color{codepurple},
    basicstyle=\ttfamily\footnotesize,
    breaklines=true,                 
    captionpos=b,                    
    numbers=left,                    
    numbersep=5pt,                  
    showspaces=false,                
    showstringspaces=false,
    showtabs=false,                  
    tabsize=4,
    language=Python
}

\usepackage[textsize=tiny]{todonotes}

\usepackage[preprint]{neurips_2026}

\usepackage[utf8]{inputenc} %
\usepackage[T1]{fontenc}    %
\usepackage{hyperref}       %
\usepackage{cleveref}
\usepackage{url}            %
\usepackage{booktabs}       %
\usepackage{amsfonts}       %
\usepackage{nicefrac}       %
\usepackage{microtype}      %
\usepackage{xcolor}         %
\usepackage{makecell}
\usepackage{mdframed}
\title{Controllably Efficient Language Models}

\author{%
  Jatin Prakash \quad
  Aahlad Puli \quad
  Rajesh Ranganath \\
  New York University \\
  \texttt{jatin.prakash@nyu.edu}
}

\begin{document}

\maketitle

\input{sections/000_abstract}

\input{sections/00_intro}

\input{sections/01_method}

\input{sections/04_related_work}

\input{sections/02_results}

\input{sections/03_discussion}
\input{sections/05_conclusion}

\bibliographystyle{plainnat}
\bibliography{paper}

\newpage

\appendix

\input{sections/06_appendix}

\end{document}

%% file: preamble/math_commands.tex
\newtheorem*{assumption*}{Assumption}
\newtheorem*{corollary*}{Corollary}
\newtheorem*{definition*}{Definition}
\newtheorem*{lemma*}{Lemma}
\newtheorem*{proposition*}{Proposition}
\newtheorem*{theorem*}{Theorem}

\renewcommand{\mid}{~\vert~}

\newcommand{\mbc}{\mathbf{c}}

\newcommand{\mbx}{\mathbf{x}}

%% file: sections/000_abstract.tex
\begin{abstract}
The substantial inference costs of attention in transformers motivated the development of efficient sequence mixers: namely sparse and sliding window attention, convolutions and linear attention.
Although these approaches result in impressive reductions in inference costs, they often trade-off with quality, specifically in-context recall.
Apriori fixing this quality-cost tradeoff at training time means being suboptimal from the get-go: some downstream applications might fundamentally require more memory for in-context recall, while other tasks may require lower latency and memory.

We propose a conceptually simple  \textit{meta}-sequence mixer with inference-cost controllability: the Compress \& Attend Transformer (\cat). 
\cat decodes chunks of tokens by \textit{attending} to compressed chunks of the sequence so far.
Both compression and decoding can use any existing sequence mixer.
Decoding from the compressed sequence yields compute and memory savings, with chunk size setting the operating point on the quality-cost trade-off. Importantly, training \cat across multiple chunk sizes at once unlocks test-time control of this trade-off without any retraining, all in a single model. 

Instantiated with the most basic choice, dense attention as the mixer, \cat \textit{surprisingly} suffices to match 10 popular and diverse efficient models (linear, hybrids, sparse) on real-world long-context recall at comparable inference costs, all from a single trained model. \cat further performs competitively on long-context understanding benchmarks while providing $1.4-3.7\times$ higher generation throughput than a dense transformer.

Play with \cats at: 
\faGithub~ \texttt{\href{https://github.com/rajesh-lab/cat-transformer}{rajesh-lab/cat-transformer}} 

or at: 
\faGithub~ \texttt{\href{https://github.com/fla-org/flash-linear-attention}{fla-org/flash-linear-attention}}

Pretrained \cats looking for adoption at: \raisebox{-0.2\height}{\includegraphics[height=1em]{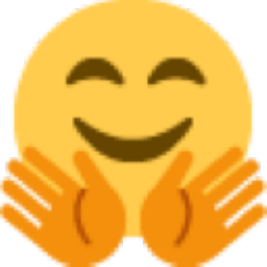}}~ \texttt{\href{https://huggingface.co/collections/bicycleman15/cat-transformer}{bicycleman15/cat-transformer}}
\end{abstract}

%% file: sections/00_intro.tex
\section{Introduction}

\begin{figure*}[t]
  \centering
  \begin{subfigure}[b]{0.48\textwidth}
    \centering
    \includegraphics[width=\linewidth]{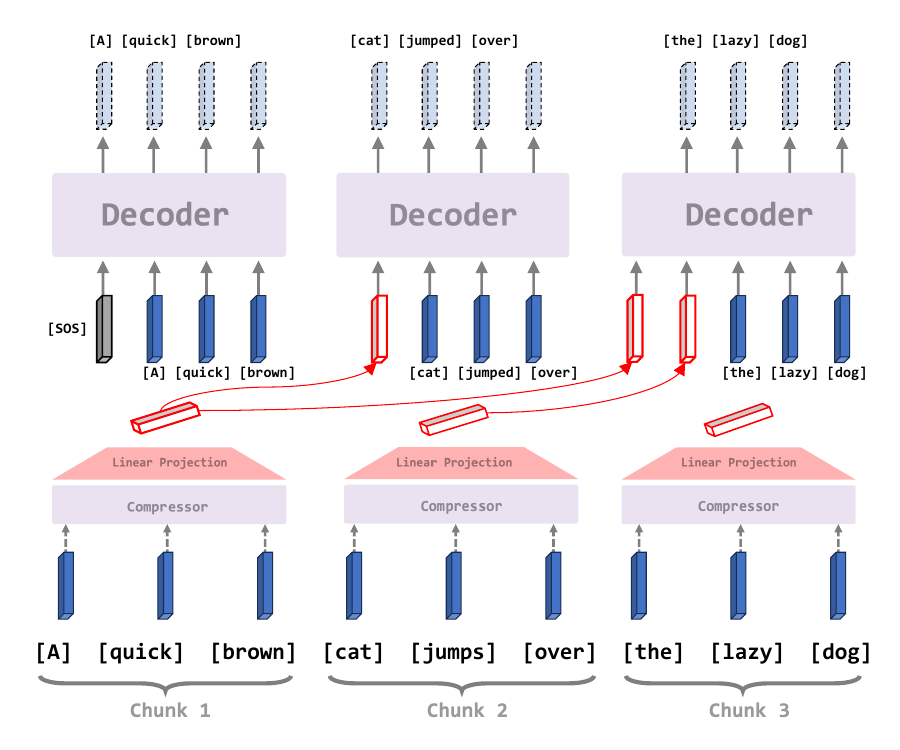}
    \caption{\textbf{The Compress and Attend Transformer (\textsc{cat}) architecture.}
    \cat chunks up a sequence of length $N$ into $N/C$ chunks of $C$ tokens (illustrated for $C=3$).
    Each chunk is parallelly compressed into a chunk representation.
    \cat then decodes each chunk by attending to past chunk representations.
    Observe the \textbf{reduced} sequence length in the decoder due to compression. The number of compressed chunks \textbf{grow} with sequence length resulting in \textit{gracefully} growing memory.
    Chunk size in \cat acts as a \textbf{\textit{knob}}, offering test-time control of quality-efficiency trade-offs, where higher chunk sizes result in more efficiency.}
    \label{fig:cat_figure}
  \end{subfigure}
  \hfill
  \begin{subfigure}[b]{0.48\textwidth}
    \centering
    \includegraphics[width=\linewidth]{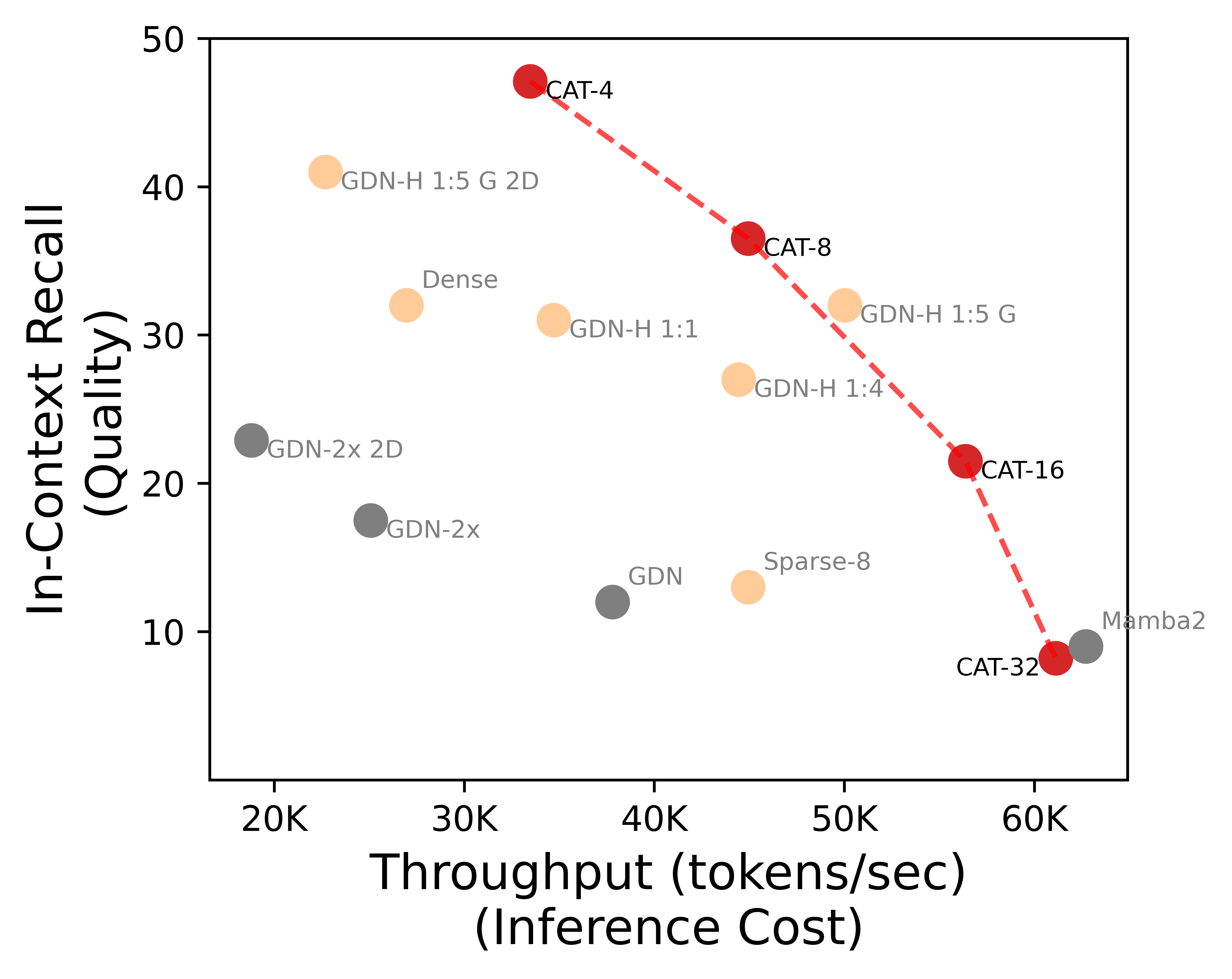}
    \caption{
    \small{
    \textbf{
    \cat unlocks test-time control of quality-inference cost trade-offs}: a \textbf{single} adaptive \cat model (\textbf{\textcolor{red}{red}} dots) matches different families of efficient approaches at comparable throughputs (inference costs) on real-world in-context recall tasks.
    We compare \textbf{across 10 models}, all having diverse model configurations, parameter counts ($\sim$300M to $\sim$820M) and \textbf{varying inference costs} for fair and broad evaluation. \textbf{\textcolor{gray}{Gray}} signifies strict competitors (linear models) where \cat cannot be used as a meta-sequence mixer, and \textbf{\textcolor{orange}{orange}} signify complementary models to \cat i.e. sequence mixers with length dependent costs (see \Cref{sec:experiments}).
}
    }
    \label{fig:pareto_frontier}
  \end{subfigure}
  \caption{Overview of \cat.}
  \label{fig:two_side_by_side}
\end{figure*}

The cost of serving language models, especially in the era of reasoning and agents, outstrips their training cost. In fact, training costs can break even with inference cost in a matter of \textit{weeks}\footnote{\texttt{together.ai} serves 400T tokens/month for open-source models (\href{https://x.com/tri_dao/status/2072429334758121556}{source}). One of the popular open-source series \texttt{Qwen-3} was pretrained only on a total of 36T tokens in total}~\citep{timbers2023llms}.
In turn, what matters now for long-term deployment 
is the quality per unit of inference cost that a language model provides.
While a model using dense attention \citep{bahdanau2014neural,vaswani2017attention} as a sequence mixer
provides good quality, its inference cost grows quickly with sequence length, making it expensive to deploy at long-contexts with the current hardware.

This large cost of dense attention at long-contexts motivated efficient alternatives in the community.
Approaches such as sparse and sliding window attention \citep{child2019generating, zaheer2020big, jiang2023mistral7b} heuristically restrict the tokens being attended to, and those such as linear attention and state-space models \citep{katharopoulos2020transformers, arora2024simple, dao2024transformers, yang2025gated} use a fixed-size recurrent state to reduce inference costs.

In all of these approaches, however, the reduced inference cost comes at the cost of quality: specifically, it trade-offs with in-context recall \citep{arora2024simple, jelassi2024repeat, wen2024rnnstransformersyetkey}. Importantly, these approaches expose this trade-off only through training-time choices. 
However, different downstream tasks have different quality requirements at test time, which a apriori training-time choice cannot accommodate. 
A general chat assistant writing short email replies has weak recall demands, and linear attention suffices; whereas a coding agent, by contrast, must recall function names across a repository-scale context, where the stronger recall of dense attention is worth its higher cost. 
A single high-recall model wastes compute on the email task, while a single low-recall model fails the coding agent's quality bar.
One way forward is to pretrain different models for different tasks, however, this gets prohibitive for all possible downstream tasks.

Hence, a single model with a knob to control the trade-off at test time becomes desirable. We note that no other existing sequence mixer \textit{natively} provides such controllability.

This paper provides a simple recipe to make inference cost controllable at test time. 
The recipe's key ingredient is a \textit{meta}-sequence mixer: a conceptually simple arrangement of existing off-the-shelf sequence mixers where one mixer compresses the sequence, and another attends to this compressed sequence while decoding.
We term this approach: \textbf{C}ompress \& \textbf{A}ttend \textbf{T}ransformer (\cat).
Concretely, \cat compresses chunks of tokens in parallel into a shorter sequence using a compressor, which a decoder then attends to while autoregressively modeling the tokens in the latest chunk.
The compression and decoding is parallel over tokens during training,
meaning there is no recurrence along the sequence (unlike \cite{Rae2020Compressive}), 
enabling \textit{end-to-end scalable training}.
When the cost of the sequence mixer used in \cat depends on the length of the sequence, decoding happens at a reduced sequence length due to compression (see \Cref{fig:cat_figure}) enabling \textit{compute and total memory savings}.
Importantly, training \cat across multiple chunk sizes at once \textbf{unlocks controllability of quality-inference cost trade-offs directly at test-time}.

We implement \cat{} with the most vanilla off-the-shelf components: compression and decoding both use dense transformer black-boxes (see \Cref{fig:cat_figure}). 
Since dense attention is a staple and well-developed, training and inference can be done efficiently with existing infrastructure.
While one can instantiate the \cat meta-sequence mixer with other mixers (say hybrids, see \Cref{tab:cat_complementary}), we find \cat with simple dense attention is already \textit{surprisingly} effective: across benchmarks, \cat matches or surpasses several popular alternatives without custom kernels or requiring careful choices (say choosing attention-linear ratios in hybrids \cite{wang2025systematic}) across inference budgets.
Notably, this is achieved with a \textbf{single} \cat model whose inference cost can be controlled on the fly.
    
Finally, comparing quality based on different architectural traits, such as parameters, FLOPs, and memory footprint\footnote{an extreme example is: imagine an architecture that has constant memory but keeps doing repeated reads and writes to this constant memory. While this has constant memory, the latency is poor due to high memory bandwidth in current hardware.}, \textit{may} be misleading: two models with identical FLOPs or parameters can differ substantially in the actual monetary cost to serve them due to 
poor implementations and hardware \textit{unfriendly} design. 
What matters is the hardware cost you actually pay for.
Hardware gets priced in dollars per hour, which means the quantity to measure a model by so that it translates to cost is: tokens per hour (division gives the cost per token) or alternatively tokens per second, both of which measure throughput.\footnote{we measure throughput as the maximum tokens per second a model can achieve under fixed hardware across batch sizes.} Thus, we plot performance against throughput (see \Cref{fig:pareto_frontier}).

To summarize, this paper makes the following contributions:
\begin{itemize}
    \item 
        \textbf{A controllably efficient meta-sequence mixer.} We introduce \cat, in which a single test-time knob (chunk size) traverses the quality–cost trade-off without retraining.

    \item \textbf{A simple instantiation that matches different families of efficient alternatives at various throughput levels.} Instantiating \cat with vanilla dense attention yields a single adaptive model that:
    \begin{itemize}

        \item achieves strong quality–throughput trade-offs compared to 10 efficient models on real-world long-context recall, without careful choices or coding-up custom kernels (\Cref{fig:pareto_frontier}).

        \item is competitive in long-context modeling and understanding benchmarks (\Cref{tab:lm_eval})
        
        \item matches the dense transformer on language modeling and is upto $1.4-3.7\times$ faster on throughput depending on the chosen chunk size (\Cref{fig:generation_throughput})
        
        \item surpasses dense transformer on real-world recall tasks using the most accurate setting (\cat-4) while still providing better throughputs ($1.45\times$) and is $1.5\times$ faster and $2\times$ memory efficient; \Cref{fig:pareto_frontier}.
    \end{itemize}
        
    \item
      \textbf{Off-the-shelf implementation.} We provide a parallel, scalable training implementation (scaling from 90M to 1B parameters) and a pure-PyTorch generation implementation requiring no custom CUDA or Triton kernels -- unlike most efficient alternatives.

    \item \textbf{Extensibility.} \cat as a meta-sequence mixer wraps around existing sequence mixers (e.g., hybrids) and can serve as a drop-in replacement layer in other architectures (\Cref{tab:cat_complementary}).
\end{itemize}

%% file: sections/01_method.tex
\section{\textbf{C}ompress and \textbf{A}ttend \textbf{T}ransformers (\cats)}
\label{sec:cats}

\paragraph{Compression and decoding.} 
\cat is a meta–sequence mixer that uses one sequence mixer to compress chunks of a sequence and another to decode within each chunk given compressed representations. 

Concretely, given a sequence $\mbx=(x_1, x_2, \dots, x_N)$ of $N$ tokens, we split the sequence into chunks $(\mbc_1, \mbc_2, \dots, \mbc_{N_C})$ containing $C$ tokens each, such that $\mbc_i = (x_{C \cdot i+1}, \dots, x_{C \cdot i + C})=(\mbx_{i,1}, \dots \mbx_{i,C})=\mbx_{i,:}$, where $\mbx_{i,:}$ indexes the $i$-th chunk of $C$ consecutive tokens (\texttt{numpy} array slicing).
\cat compresses each chunk $\mbc_i$ using the {compressor} $f_\theta$ into chunk representations. The {compressor} $f_\theta$ %
is any sequence mixer with hidden size $D_{f}$, followed by a linear projection to $D_g$. This leads to a {compressed} chunk representation $f_\theta(\mbc_i)\in \mathcal{R}^{D_g}$. 
That is:
\[
\{x_1, \cdots x_N\} \;\xrightarrow{\text{chunk}}\; \{\mbc_i\}_{i=1}^{N_c} \;\xrightarrow{\text{compress}}\; \{f_\theta(\mbc_i)\}_{i=1}^{N_c}
\]
After compression, \cat decodes the original sequence $\mbx$ from the compressed chunk representations $\{f_\theta(\mbc_i)\}_{i=1}^{N_C}$ using a {decoder} $g_\theta$, which is a causal sequence mixer, having hidden size $D_g$, matching the linear projection from the compressor.
\cat decodes chunks autoregressively, where to decode each token $\mbx_{i,j}$ in a chunk $\mbc_i$, the decoder takes as input the previous tokens $\{\mbx_{i,<j}\}$ in chunk $\mbc_i$ and the past chunk representations $\{f_\theta(\mbc_1), \dots, f_\theta(\mbc_{i-1})\}$. Formally, the predictive distribution $p_\theta$ for the tokens in chunk $\mbc_i$ is defined as:
\begin{equation}
\begin{multlined}
p_\theta(\mbc_i \mid \mbc_{i-1}\cdots \mbc_1) =
\prod_{j=1}^C
g_\theta\!\left(
  \underbrace{\mbx_{i,j}}_{j^{\text{th}}\ \text{token in chunk }\mbc_i}
  \,\middle|\,
  \substack{
    \underbrace{\mbx_{i,j-1},\dots,\mbx_{i,1},}_{\text{previous tokens in chunk }\mbc_i}
    \underbrace{f_\theta(\mbc_{i-1})\cdots f_\theta(\mbc_1)}_{\text{past chunk representations}}
  }
\right)
\end{multlined}
\end{equation}

When decoding cost depends on sequence length, \cat improves throughput by reducing the compute and memory required, using compressed chunk representations; the larger the chunk size, the larger the reduction.

During training, compression and decoding happen in parallel for all tokens in the sequence because the compression of a chunk does not depend on earlier chunks.
This choice allows the entire \cat model to be efficiently trained end-to-end with the standard next-token prediction loss.
The end-to-end training lets \cat \textit{learn what to retain} in its compressed chunk representations, rather than relying on fixed attention patterns or complex state update rules.

\paragraph{Training for test-time control of inference cost.}
\label{sec:adaptive_cats}
Changing the chunk size in \cat trades off quality for compute and memory efficiency.
Training \cat with multiple chunk sizes yields a single adaptive model whose compute-memory budget can be adjusted at test time without retraining.
We uniformly sample a chunk size $C$ at each training iteration and pass a \textit{learnable} indicator token to \cat to indicate the current chunk size.
The compressed tokens are separated from the uncompressed ones in the decoder using a marker token shared across chunk sizes.
After training, one can use the same \cat model at different compute/memory budgets by changing the indicator token at test time.
\Cref{app:adaptive_cat_train_details} provides further details.

\textbf{\cat as a layer.}
The principles discussed above are architecture-agnostic: \cat can be instantiated as a modular layer that can be inserted in any architecture, unlocking controllable cost there and enabling new hybrid designs.
In layer form, a simple linear projection can serve as the compressor, and a sequence mixer (e.g., dense attention) as the decoder.
\Cref{app:cat_as_layer} provides preliminary results; full exploration is left to future work.

\subsection{How to implement fast and scalable \cats}
\label{sec:implement_cats}

\begin{tcolorbox}[
    colback=white,
    colframe=gray!50,
    boxrule=0.5pt,
    arc=2pt,
    left=6pt, right=6pt, top=4pt, bottom=4pt,
]
For simplicity, we instantiate the sequence mixers for both compression and decoding as \textit{vanilla dense attention}, demonstrating just how \textbf{far} such simple design choices can go with \cat, yielding a competitive and test-time controllable architecture.
With dense transformers serving as both the compressor and the decoder, \cat admits a pure PyTorch implementation for scalable training and fast generation requiring no custom CUDA or Triton kernels. We outline this approach below.
\end{tcolorbox}

\textbf{Fast and Parallel Compression.} Compression of chunks of tokens is efficient and can be executed in parallel, for instance by using \texttt{torch.vmap}, to produce $\{f_\theta(\mbc_i)\}$ for all chunks $\mbc_i$. This costs a total of $O(\frac{N}{C}\cdot C^2)=O(NC)$ in self-attention compute, rather than $O(N^2)$.

\textbf{Naive and Slow Training.} For training the decoder, a naive implementation can lead to slower training. To compute logits for tokens in chunk $\mbc_i$, that is computing $g_\theta (\mbc_i \mid f_\theta(\mbc_1) \cdots f_\theta(\mbc_{i-1}))$ in parallel can be non-trivial. Since, for chunk $\mbc_i$, the number of past chunks varies, making shapes variable and as a result, harder to parallelize the computation of logits.
One could employ a python loop and compute logits for every chunk sequentially, but that would be slow and would not scale.
Padding to make shapes constant to allow parallelism would make things worse by increasing wasteful computations.
In fact, even if one bypasses varying shapes problem and manages to compute logits for every chunk in parallel, the total self-attention operations in the decoder would scale as $O(\sum_{i=1}^{N_c}(i+C)^2)=O((\frac{N}{C})^3)$, that is cubic in sequence length.
Thus, even the ideal parallel approach for training will not scale, despite the simplicity of \cat.

\paragraph{Parallel and Scalable Training.} To overcome above \textbf{training challenges} in \cats, we observe that in computing logits for every chunk $\mbc_i$, one calculates exactly the same key-value vectors for the representation $f_\theta(\mbc_j)$ in the decoder transformer, where $j<i$.
This means that computation is duplicated.
We exploit this observation in training \cats.
We implement training by interleaving compressed representations into the sequence: $\{\mbc_1, f_\theta(\mbc_1), \mbc_2, f_\theta(\mbc_2), \dots\}$. A custom attention mask (App. Figure~\ref{fig:cat_attention_mask}) lets a token in chunk $\mbc_i$ attend to earlier tokens in the same chunk and to prior chunk representations $f_\theta(\mbc_{<i})$, but not to raw tokens in other chunks. This lets the decoder reuse the keys and values of $f_\theta(\mbc_i)$ when computing logits for any later chunk. The resulting complexity is $O(N^2/C)$ -- a constant-factor improvement over the dense transformer's $O(N^2)$, enabling potentially faster pre-training (see \Cref{sec:discussion}).

\textbf{Fast and Efficient Generation.}
Due to compression, \cats can throwaway past chunks of tokens, and only keep their compressed chunk representations in memory.
This straightaway results in a big reduction of memory; the KV cache is slashed by a factor of $C$, even for a modest chunk size of 4 (see \Cref{fig:generation_throughput}).
\textbf{Notably these memory savings are \textit{independent} of sequence length}; in other words, \cat always results in memory reduction \textbf{\textit{relative}} to a dense transformer at any sequence length (be it 4K or 128K).
This compressed sequences means both reduced HBM accesses and that the decoder attends to atmost $\frac{N}{C}+C$ tokens during generation, resulting in an increased throughput.

Implementing generation is similar to how it occurs for a dense transformer. A pure PyTorch implementation\footnote{Our implementation is inspired from: \href{https://github.com/meta-pytorch/gpt-fast}{github.com/meta-pytorch/gpt-fast}} for \cats is on-par with efficient architectures that utilize custom kernels.
Given a prompt, \cat first computes chunk representations in parallel and prefills them into the decoder's KV cache. Generation then proceeds chunk by chunk: tokens within a chunk are decoded sequentially, and on chunk completion the chunk is compressed and its representation is appended to the KV cache before decoding continues.
Further details and a PyTorch style pseudo-code are in \Cref{app:pseudo_code,app:generation_details}.

\subsection{Scaling model parameters without proportionally increasing inference costs.}
\label{sec:cat_use_more_parameters}
The decoder holds the majority of parameters in \cat, and dominates overall inference cost (Section~\ref{sec:experiments}).
However, because it operates on a compressed sequence rather than the full one, its compute and memory requirements are dramatically lower than a dense transformer, at the same parameter count.
With this saved budget, we use a larger decoder in \cat: scaling parameters while keeping inference cost below that of the smaller parameter dense transformer.
The result is an improved trade-off (\textbf{\textcolor{blue}{blue}} $\rightarrow$ \textbf{\textcolor{red}{red}}, see appendix \Cref{fig:parameter_matched_figure}).

While adding parameters improves quality in any architecture, whether the added cost is justified depends on the architecture itself.
Increasing parameters in a linear model (\textcolor{gray}{\textsc{gdn-2$\times$} $\rightarrow $ \textsc{gdn-2$\times$ 2D}}) to parameter match \cat improves quality, but disproportionately increases costs (decreases throughput) (\Cref{fig:pareto_frontier}), yielding a worse quality-throughput trade-off than \cat (\textbf{\textcolor{red}{red}} dots). 
Similar results holds for other models (\textcolor{gray}{Sparse-8, \textsc{gdn-h 1:5 G 2D}}) which we discuss in \Cref{sec:experiments}.
Thus, the \cat computational structure, or more generally, the model architecture \textit{matters}~\footnote{As an extreme example, an embedding-plus-unembedding model with an arbitrarily large embedding dimension has many parameters but can only represent token bigrams.} beyond the parameter count.
We believe the \textit{careful} design of the \cat meta-sequence mixer: gracefully growing memory and reduced sequence length due to compression contribute to this better quality-inference cost trade-off despite increased parameters in the decoder.
Further, we note this decoupling of parameter count from inference cost to increase quality-cost trade-off is analogous to MoEs\footnote{Note that while MoEs do enable scaling of parameters while controlling costs, they are not sequence mixers since they operate on the feedforward layers and can be applied to any mixer, including \cats.}~\citep{shazeer2017outrageously}

Most sequence mixers and architectures are \textit{monolithic} and incur relatively higher inference costs when increasing parameters.
From a deployment perspective, what matters is performance at a given inference cost: 
if \cats can deliver better performance at the same cost (better trade-off), owing to their additional parameters \textbf{and} computational structure, this is a desirable property.
To motivate this \textbf{inference-first design of} \cat further, we ask a question: \textit{suppose model A has more parameters than model B, then if model A outperforms model B using lower inference costs, does it matter that model A has more parameters than model B? Which model should one deploy: model A or B?}

For completeness, we provide results when \cat{} is parameter-matched to few of the lower-parameter models in our comparison at \Cref{app:parameter_matched}.
We observe parameter-matched \cat{} achieve the lowest inference cost among most models considered in our evaluation, and as a result are \textit{incomparable}.

%% file: sections/04_related_work.tex
\begin{table*}[h]
\scriptsize   %
\setlength{\tabcolsep}{3pt}
\renewcommand{\arraystretch}{1.3}
\begin{tabular}{|p{0.18\linewidth}|p{0.10\linewidth}|p{0.09\linewidth}|p{0.09\linewidth}|p{0.14\linewidth}|p{0.09\linewidth}|p{0.10\linewidth}|p{0.10\linewidth}|}
\hline
\textbf{Method} & \textbf{Unrestricted Access to Memory?} & \textbf{Flexible memory?} & \textbf{Scalable training?} & \textbf{Both compute \& memory efficient?} & \textbf{Controllable inference costs?} & \textbf{Mixable with \cat?} & \textbf{Usable in \cat's decoder?} \\
\hline
\textbf{\textit{Dense}}: \cite{vaswani2017attention}
  & \cmark{} %
  & \cmark
  & \cmark
  & \xmark
  & \xmark
  & \cmark
  & \cmark \\
\hline
\textbf{\textit{Sparse Attention}}: \cite{child2019generating}
  & \xmark{} %
  & \cmark
  & \cmark
  & \cmark
  & \xmark
  & \cmark
  & \cmark \\
\hline
\textbf{\textit{NSA}}: \cite{yuan2025native}
  & \cmark{} %
  & \cmark
  & \cmark
  & \xmark
  & \xmark
  & \cmark
  & \cmark \\
\hline
\textbf{\textit{Sliding window Attn.}}: \cite{jiang2023mistral7b}
  & \xmark
  & \xmark
  & \cmark
  & \cmark
  & \xmark
  & \cmark
  & \cmark \\
\hline
\textbf{\textit{Linear Attention}:} \cite{dao2024transformers}
  & \cmark{} %
  & \xmark
  & \cmark
  & \cmark
  & \xmark
  & \cmark
  & \xmark \\
\hline
\textbf{\textit{Recursive compression}}: \cite{chevalier2023adapting}
  & \cmark
  & \cmark
  & \xmark{} %
  & \cmark
  & \xmark
  & \cmark
  & \cmark \\
\hline
\textbf{\textit{MegaByte/Block Transformer}}: \cite{ho2024block, yu2023megabyte}
  & \cmark
  & \xmark
  & \cmark
  & \cmark
  & \xmark
  & \cmark
  & \cmark \\
\hline
\textbf{\textit{\cats}}
  & \cmark %
  & \cmark %
  & \cmark %
  & \cmark
  & \cmark
  & \cmark
  & \cmark \\
\hline
\end{tabular}
\caption{
We categorize existing related work by key properties desirable for an efficient architecture, and indicate whether \cat{} can complement these approaches as a \textbf{meta-sequence mixer}.
\textit{``Both compute and memory efficient?''} signifies savings during inference;
\textit{``Unrestricted Access to Memory''} signifies whether an architecture can freely access any part of the memory in the past, without any artificial restrictions;
\textit{``Mixable with \cat?''} indicates whether \cat{} as a layer be used in these approaches;
\textit{``Usable \cat's compressor/decoder?''} indicates whether the method can serve as a compressor or decoder within \cat. Note that \cat itself can be recursively used as compressor/decoder.
}
\label{tab:related_work}
\end{table*}

\section{Related work.}
Efficient sequence mixers reduce attention's cost in different ways: sparse and sliding window attention restrict which tokens are attended to \citep{child2019generating, zaheer2020big, jiang2023mistral7b}; linear attention and state-space models replace softmax with a fixed-size recurrent state \citep{katharopoulos2020transformers, dao2024transformers, yang2025gated}; recurrent compression accumulates state sequentially \citep{Rae2020Compressive, chevalier2023adapting}; and hierarchical or chunk-based architectures compress the sequence into coarser units \citep{nawrot2021hierarchical, yu2023megabyte, ho2024block, pagnoni2025byte}. Each family trades off something — recall \citep{arora2024simple, jelassi2024repeat, wen2024rnnstransformersyetkey}, training scalability \citep{geiping2025scaling}, or careful hybrid tuning \citep{waleffe2024empirical, wang2025systematic} — and crucially, \textbf{none expose test-time control of inference cost}. \cat is complementary rather than competing: any length-dependent mixer can serve as its compressor or decoder and inherit controllability, and orthogonal techniques (e.g., MoEs \citep{shazeer2017outrageously}, speculative decoding, training-free sparsification \citep{nawrot2025sparse}) compose on top. 

In summary, \cat complements most existing (or future) approaches, can extend them, or be mixed with them to unlock test-time control of inference cost.
\Cref{tab:related_work} highlights the relationships and conceptual differences between \cat and the most relevant related work.
App. \ref{app:extended_related_work} provides the full related work.

%% file: sections/02_results.tex
\section{Experiments}
\label{sec:experiments}

\subsection{Models in Comparison and Training Setup}

\cat{} as a \textit{meta}-sequence mixer complements any sequence mixer whose decoding cost depends on sequence length -- instantiating \cat{}'s decoder with such a mixer reduces costs due to the compressed sequence. 
This includes sequence mixers like dense attention including hybrid models using attention in few layers.
Consequently, the only \textit{strict competitors} are sequence mixers with length-independent costs, such as pure linear attention models (which cannot be composed with \cat; see \Cref{tab:related_work}).
The main results reflect this categorization: into (a) \textcolor{gray}{strict baselines} and (b) \textcolor{orange}{\textit{complementary to \cat}}, for clarity.
That being said, we compare broadly against recent state-of-the-art architectures (across hyper-parameters) of both kinds with different quality-throughput trade-offs, and show that a single \cat model with simple dense attention, to our surprise, is \textbf{highly competitive across the throughput levels}.

\textbf{Models in comparison:} 
Our evaluations include: (i) attention-based: standard dense transformer \citep{touvron2023llama} and sparse transformer \citep{child2019generating}, (ii) Linear Transformers such as Mamba2 \citep{dao2024transformers} and GatedDeltaNet (\textsc{gdn}) \citep{yang2025gated}, as well as (iii) Hybrid architectures with \textsc{gdn} and attention layers interleaved in some pre-specified ratio.
By default, all models below are configured with
$L=12$ layers and $D=1024$ hidden dimension; any deviations are explicitly stated below.

Further, going beyond the default hyperparameter settings, we scale up each model type by adjusting their model configurations to yield models at varying throughput levels: (i) linear attention with a $2\times$ recurrent state size (i.e. \textsc{gdn}-2$\times$), (ii) multiple dense-linear attention ratios in hybrid architectures (i.e. \textsc{gdn-h 1:1}, \textsc{gdn-h 1:4}), (iii) hybrids with global attention layers (i.e. \textsc{gdn-h 1:5 G}), (iv) and $2\times$ the model dimension (for instance \textsc{gdn $2\times$ 2D}, \textsc{gdn-h 1:5 G 2D}, Sparse-4/8). 
Most model types has atleast $2$ configurations to ensure fair and broader comparison.
This process results in a total of 10 models with parameter counts ranging from $\sim$250M \textbf{to upto $\sim$820M}, and \textbf{varying} throughput levels.

\textbf{We compare these 10 models against a \textit{single} \cat model.} 
The only \textcolor{gray}{strict baselines} to \cat are sequence mixers whose inference costs are length independent -- this includes Mamba2, \textsc{gdn}, \textsc{gdn}-$2\times$ and the larger parameter \textsc{gdn}-$2\times$ 2D. 
The rest in our broad comparison are \textcolor{orange}{\textit{complementary to \cat}}.
See~\cref{sec:discussion} for further discussion on the complementary nature of \cats (\Cref{tab:cat_complementary}).

\textbf{\cat configuration:} For \cat, we use $L=12$ layers (same as baselines), and a wider hidden size of $D_g=2D=2048$ for the decoder, that takes up the majority of the parameters. 
The compressor is small and uses $L=3$ layers and hidden size of $D_f=D=1024$. 
Depth of compressor does not have major effect (App. \ref{app:cat_ablation}).
This makes the parameter count for \cats close to $\sim(820+150)$ M parameters (similar to some models included in our comparison).
We train \cat simultaneously on chunk sizes $C=\{4,8,16,32\}$.
This yields a single model that can work with different chunk sizes at once, offering test-time control of inference costs.

\textbf{Setup:} All models were trained on 15B tokens of FineWeb-Edu \cite{penedo2024fineweb} with a context length of 4K following \cite{behrouz2024titans, yang2025gated}.
We use the AdamW optimizer \cite{loshchilov2017decoupled} with a peak learning rate of 8e-4, weight decay of 0.1, gradient clipping of 1.0, batch-size of 0.5M tokens, employing the GPT2 tokenizer.
App. \ref{app:training_details} provides details for each model and training. To obtain throughput, refer to details at \Cref{app:generation_details}.

\begin{figure}[ht]
    \centering
    \hspace{-30pt}\includegraphics[width=0.92\linewidth]{figures/main_results_figure_v2.png}
    \caption{
    We report quality-throughput trade-offs for different models across diverse long-context tasks.
    \cat provides a strong trade-off, all in a single model.
    Refer to
    \Cref{tab:s_niah_main} for needle-in-haystack task (NIAH-N);
    \Cref{tab:longppl_longbench} in appendix reports full evaluations on the LongPPL and LongBench.
    }
    \label{fig:main_results_figure}
\end{figure}

\subsubsection{Results}

\textbf{Long-context recall:}
\Cref{fig:pareto_frontier} reports results on real-world in-context recall tasks from \cite{arora2024simple} for a given throughput requirement.
We report results on SWDE and FDA, which have longer sequences among the datasets in the suite (others have an average length of $\leq300$ tokens~\citep{arora2024just}). Table \ref{tab:swde_fda_results} provides numbers, and \ref{app:recall_evaluation} shows evaluations on all datasets.
\textbf{\cat performs as good or better than all models across inference cost budgets (throughputs), using a single model only.}
Linear models (Mamba2, \textsc{gdn}) lag far behind dense attention, while hybrids reduce the gap.
\textbf{\cat provides strong recall-cost trade-off, benefiting from the gracefully growing memory}, i.e. growing linearly but by a constant factor less compared to dense attention.
\cat interestingly outperforms even the dense transformer at these tasks (at moderate chunk sizes $=4,8$), while having a higher throughput.
\Cref{app:datasets} provides more details about the task.
 
\Cref{fig:main_results_figure} reports results on the needle-in-haystack task (NIAH-N) from the RULER benchmark \cite{hsieh2024ruler}, that is, retrieve a seven token number from long-context.

\cats outperform the existing efficient models as context length increases, and \textbf{interestingly show \textit{slower} degradation with length}.
This slow degradation can possibly be attributed to reduced sequence length in \cat that leads to fewer \textit{distractions} for the saturating dense attention on long-contexts \citep{barbero2024transformers, vasylenko2025long, chiang2022overcoming, golovneva2025multi}.
More discussion can be found in App. \ref{app:niah}, where we extend these results to the harder task from RULER (namely NIAH-U, that is retrieve 32 token \texttt{uuid}s).

\textbf{Long-context language modeling and understanding:}
We test language modeling and understanding on long contexts (upto $4$K contexts). 
As standard perplexity (or test log-loss) averaged over all tokens \textbf{does not} indicate downstream long-context ability \citep{fang2025what, liu2023same, hu2024can}, \cref{fig:main_results_figure} conducts evaluations on the LongPPL \cite{fang2025what} metric, that calculates loss on a few key tokens which are essential for long-context understanding. 
We employ Llama-3.1-8B as evaluator.
Additionally, \cref{fig:main_results_figure} reports evaluation on LongBench \cite{bai2023longbench} that tests for long-context understanding.
All \cats perform competitively on longer contexts. 

\begin{wrapfigure}[13]{l}{0.5\textwidth}
    \centering
    \includegraphics[width=0.45\textwidth]{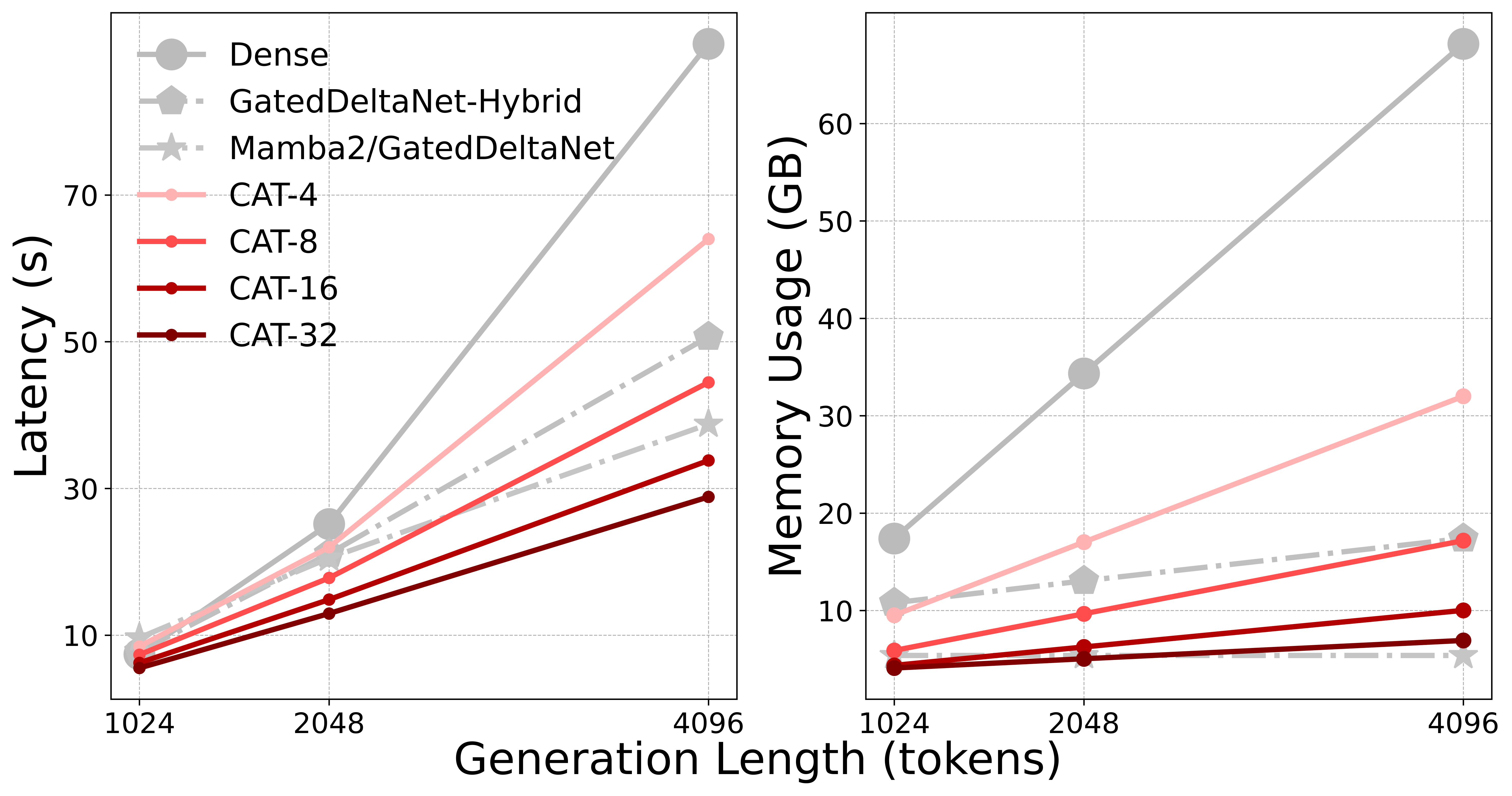}
    \captionsetup{font=footnotesize,labelfont=footnotesize}
    \caption{A single \cat model generates $1.4-3.2\times$ faster than the dense transformer while showcasing upto $2.2-9.5\times$ lower memory usage.}
    \label{fig:generation_throughput}
\end{wrapfigure}

\begin{wraptable}[7]{r}{0.5\textwidth}
\vspace{-170pt}
\footnotesize
\centering
\captionsetup{font=footnotesize} %
\caption{Accuracy on RULER \cite{hsieh2024ruler} S-NIAH-N benchmark and average LM evals \cite{eval-harness}. All \cat results come from a single model, evaluated at different chunk sizes. \Cref{fig:main_results_figure} plots these results against throughput. \Cref{tab:s_niah,tab:lm_eval} in the appendix provide more results.}
\label{tab:s_niah_main}
\setlength{\tabcolsep}{4pt}
\begin{tabular}{lccc|c}
\toprule
 & \multicolumn{3}{c|}{\textbf{S-NIAH-N} ($\uparrow$)} & \textbf{LM Evals}\footnotemark[\value{footnote}] ($\uparrow$) \\
\cmidrule(lr){2-4} \cmidrule(lr){5-5}
\textbf{Model} & \textbf{1K} & \textbf{2K} & \textbf{4K} & \textbf{Avg.} \\
\midrule
\textit{Strict baselines}           &          &          &          &          \\
\textsc{gdn}    & 84.7         & 69.1         & 13.6         & 43.5     \\
\blue{\textsc{gdn}-2$\times$}    & 78.0         & 61.4         & 29.0         & 43.8     \\
\blue{\textsc{gdn}-2$\times$ 2D}    & 99.3         & 97.0         & 72.0         & 46.4 \\
\midrule
\cat-4          & 96.0         & \ud{97.0} & \textbf{96.0} & 43.9 \\
\cat-8          & 90.0         & {93.0}     & \ud{91.0}     & 44.1 \\
\cat-16         & 76.0         & 72.0          & 70.0          & 44.3 \\
\cat-32         & 60.0         & 37.0          & 31.0          & 45.1 \\
\midrule
\textit{Complementary}           &          &          &          &          \\
Dense           & 96.0         & 92.0         & 43.0         & 42.1 \\
\textsc{gdn}-H 1:1 & \ud{99.0} & 97.0 & 44.0     & 43.0 \\
\blue{\textsc{gdn}-H 1:5 G 2D}    & \textbf{99.5}         & \textbf{99.3}         & 43.8         & 45.8 \\
\bottomrule
\end{tabular}
\end{wraptable}
\footnotetext[\value{footnote}]{common LM eval measures performance on \textbf{short sequences} ($\leq 30$ tokens on average), which is \textit{not} indicative of long-context performance, as reported multiple times in the literature \cite{bertsch2025cracks,fang2025what}. We perform the eval for completeness and transparency.}

\textbf{Short-context language modeling and understanding benchmarks:}
The last column in \Cref{tab:s_niah_main}
(and \Cref{tab:lm_eval} in appendix) reports zero-shot accuracies on key common-sense reasoning benchmarks for completeness.

\textbf{Benchmarking generation:}
Figure \ref{fig:generation_throughput} compares architectures as one scales the sequence length, with a fixed batch-size of 320 to maximize throughput.
\cat generates sequences \textbf{$1.4-3.2\times$ faster} than the dense transformer while showcasing \textbf{upto $2.2-9.5\times$ lower total memory usage} as one increases chunk sizes, despite using significantly more parameters than the baselines due to wider decoder and the additional compressor.
This is not surprising since the major bottlenecks during generation are: (a) KV cache size that drives the main memory requirement during generation and not the parameter count (Sec. \ref{sec:discussion}), (b) memory accesses required for a token, and (c) FLOPs used per token determined by the past tokens being attended to.
\cats reduce these factors despite carrying more parameters overall.
App. \ref{app:generation_details} provides implementation details.

\textbf{Appendix:} We provide additional results demonstrating \cat's flexibility and scalability. First, we show \cat can be used as a drop-in layer (App.~\ref{app:cat_as_layer}) and that its compressor and decoder can make use of any sequence mixer (App.~\ref{app:cat_as_meta_mixer}). 
We then present scaling experiments (App.~\ref{fig:scaling_law}), ablations on design choices (App.~\ref{app:cat_ablation}), evaluations on the synthetic MQAR task (App.~\ref{app:synth_task} -- tested on $4\times$ longer sequences than usual), and analysis of perplexity across chunk boundaries (App.~\ref{app:across_chunk_anal}).

\subsubsection{Long(er) Context Evaluation}

\begin{table}[t]
\begin{minipage}[t]{0.56\textwidth}
\vspace{0pt}
{\footnotesize
\centering
\renewcommand{\arraystretch}{0.95}

\begin{tabular}{lccccc}
\toprule
\textbf{Model} & \textbf{2K} & \textbf{4K} & \textbf{8K} & \textbf{16K}$^\dagger$ & \textbf{LM evals} $\uparrow$ \\
\midrule
\cat-4    & 99.0 & \ud{90.4} & \textbf{98.0} & \textbf{90.1} & 44.6 \\
\cat-8    & 94.4 & \textbf{92.0} & \ud{87.0} & \ud{82.1} & 44.5 \\
\cat-16   & 82.7 & 82.7 & 57.8 & 51.8 & 45.3 \\
\cat-32   & 71.8 & 42.9 & 26.2 & 11.6 & 45.8 \\
\midrule
Dense    & \ud{99.3} & 59.5 & 37.9 & 5.0 & 44.0 \\
\textsc{gdn-h 1:1}  & \textbf{99.7} & 53.2 & 24.9 & 13.2 & 44.4 \\
\bottomrule
\end{tabular}
\caption{
{\footnotesize Longer-context evaluation on RULER NIAH-N \cite{hsieh2024ruler}. \cat degrades \textit{gracefully} as context grows; all \cats are a single model with test-time adjustable trade-offs. $^\dagger$16K uses continued pre-training of the 8K model on 16K sequences.}
}
\label{tab:cat-memory-niah}
}
\end{minipage}\hfill
\begin{minipage}[t]{0.40\textwidth}
\vspace{0pt}
\centering
\includegraphics[width=0.85\linewidth]{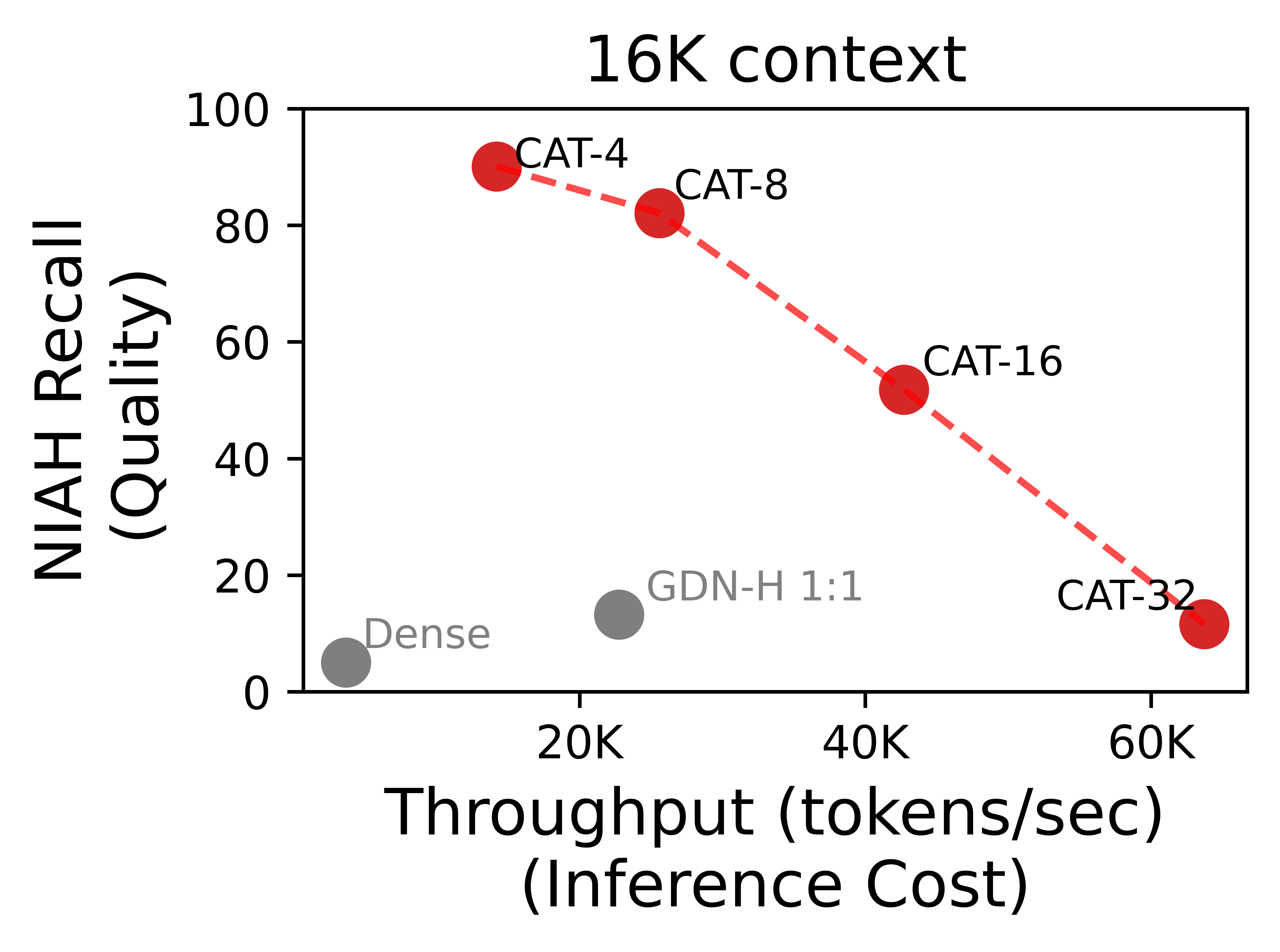}
\captionsetup{font=footnotesize,labelfont=footnotesize}
\captionof{figure}{\Cref{tab:cat-memory-niah} plotted against throughput.}
\label{fig:16k_result}
\end{minipage}
\end{table}

We provide scaled up results on 8K sequence length for a few models in our comparison and \cat.
All models were trained from scratch on 8K sequence length for 30B tokens on FineWeb-Edu \cite{penedo2024fineweb}. We additionally increased the number of layers to 18 for all models.
\Cref{tab:cat-memory-niah} reports these.
We restrict this study to a subset of models due to compute constraints.

To stress test at longer contexts, we perform continued pre-training for all models for additional 1B tokens at 16K sequence length, and then evaluate them on NIAH-N task again.

%% file: sections/03_discussion.tex
\section{Discussion}
\label{sec:discussion}

\begin{wrapfigure}[25]{r}{0.42\textwidth}
    \vspace{-12pt}
    \centering
    \includegraphics[width=\linewidth]{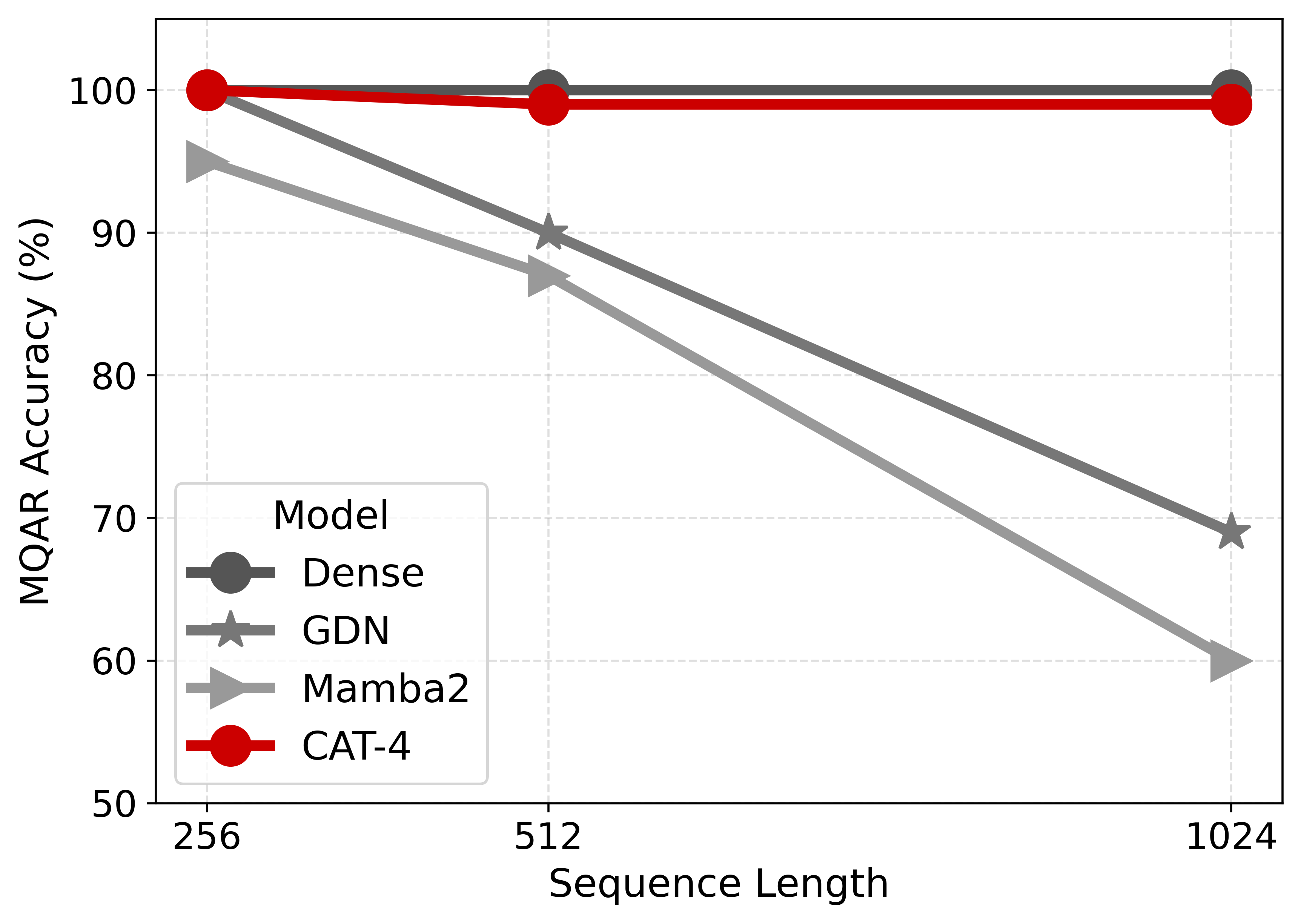}
    \caption{Comparison of sequence mixers on MQAR \cite{arora2023zoology} task (up to $4\times$ the usual length).
        All models are \textbf{memory-matched} in bytes at \textbf{every sequence length} (except dense transformer).
        \cat outperforms linear models especially at longer sequences, while still consuming same memory (or state size), and hence providing a better trade-off, possibly due to better learning in a gracefully growing memory.}
    \label{fig:mqar_main_paper}
    \vspace{-8pt}
\end{wrapfigure}

\textbf{Gracefully growing memory.} Sequence mixers like dense attention or linear recurrences take an \textit{extreme} route for long-context modeling: either the memory keeps growing \textit{aggressively} (linearly), or the memory is fixed.
\cats take a \textbf{middle-ground}, where the memory increases \textit{gracefully} by constant factor less than dense.
This flexible memory results in better recall-cost trade-off.
Interestingly, growing memory may result in \textit{better memory management} as compared to fixed memory models \cite{wen2024rnnstransformersyetkey} despite using the same memory at a particular sequence length.
We stress test this memory management on the MQAR task \cite{arora2023zoology} in \Cref{fig:mqar_main_paper} (more details in \Cref{app:synth_task}).
\cat provides strong performance-cost trade-off of in MQAR despite taking the same memory as fixed memory models. This better trade-off also echoes in our main evaluations (\Cref{fig:pareto_frontier}).

However, note that \cat fundamentally does have length-dependent costs, and a pure linear model (say, \textsc{gdn-2$\times$ 2D}) may overtake \cat in raw throughput at very long sequences.
The relevant question, however, is not throughput alone but whether that throughput comes with \textit{usable} quality. 
Constant memory may seem attractive, but it trades off severely with long-context performance, making it ineffective at the very task it was designed for.
Hybrids recover part of this trade-off by reintroducing a growing memory of dense attention, and \cat does the same through its growing set of compressed chunk representations.
Taken together, this may point to a fundamental property for modeling long contexts well: one requires a memory that grows with the context, or at least a non-trivial memory budget that the small fixed states of linear models do not provide.
Refer to \Cref{sec:conclusion} for a discussion on future work.

\textbf{What is the training cost of \cat?}
\cat reduces attention FLOPs significantly through compressed sequences, but training cost depends on both attention and feedforward layers.
At 4K sequence length, feedforward FLOPs dominate over attention \citep{scaling-book}, so \cat's training cost is $\sim2.5\times$ that of a smaller but throughput-matched dense transformer.
This overhead shrinks as sequences lengthen and attention becomes the bottleneck: at 16K it falls to $\sim1.25\times$, and at 64K it drops below the throughput-matched dense baseline despite \cat's larger parameter count.
More importantly, \textbf{at any sequence length}, \cat \textbf{amortizes} training cost by exposing multiple operating points from a single model.
Training a separate model for each operating point would cost more in total: matching the throughputs of \textsc{gdn}, \textsc{gdn-$2\times$}, \textsc{gdn-$2\times$ 2D}, and \textsc{gdn-h 1:1} with independent runs takes at least $2\times$ the training FLOPs of a single adaptive \cat, and the gap widens as pretraining moves to longer sequences.
Since inference cost dominates the total cost of building and deploying models, the ability to serve at multiple throughput levels from a single model is more valuable than the training overhead.
See \Cref{app:cat_training_throughput} for details.

\begin{table}[t]
\footnotesize
\centering
\captionsetup{font=footnotesize,width=0.9\linewidth}
\caption{\cat composes with existing efficient sequence mixers. Using \textsc{gdn}-H 1:1 as the \cat decoder yields larger memory savings than either component alone, and even improves recall --- highlighting their complementary nature. Trained at 1K context for 5B tokens with chunk size 16. See \Cref{app:cat_as_meta_mixer} for details.}
\label{tab:cat_complementary}
\begin{tabular}{lcc}
\toprule
\textbf{Model} & \textbf{Mem.\ Savings} ($\uparrow$) & \textbf{SWDE recall} ($\uparrow$) \\
\midrule
Dense                          & $1.0\times$  & 39.2 \\
\textsc{gdn}-H 1:1             & $2.5\times$  & 28.0 \\
\midrule
\cat-16 (Dense)                & $7\times$    & 13.5 \\
\cat-16 (\textsc{gdn}-H 1:1)   & $\mathbf{14\times}$ & 16.5 \\
\bottomrule
\end{tabular}
\end{table}

\textbf{\cat is a \textit{meta}-sequence mixer.}
\cat is not a sequence mixer in the conventional sense but a \textit{meta}-architecture that wraps around any sequence mixer.
Whenever the wrapped mixer has length-dependent decoding cost --- dense, sparse, or hybrid --- \cat provides compute and memory savings via the reduced sequence length, and, more importantly, exposes controllable efficiency on top.
This makes existing and future length-dependent mixers, and any techniques that accelerate them (e.g., speculative decoding \citep{leviathan2023fast}), complementary rather than competing with \cat.
\Cref{tab:cat_complementary} reports preliminary results.
We deliberately instantiated \cat with the most basic and ubiquitous mixer --- dense attention --- to make these techniques immediately accessible.

%% file: sections/05_conclusion.tex
\section{Conclusion and Future Work}
\label{sec:conclusion}

This paper set out to provide a simple recipe for making per-token inference cost controllable at test time.
\cat delivers on this: a single \cat model exposes a test-time knob (chunk size) that traverses the quality--throughput frontier without retraining. Instantiated with plain dense attention --- an off-the-shelf component requiring no hyperparameter tuning or custom kernels --- it matches or surpasses many popular efficient baselines at various throughput levels.

\begin{tcolorbox}[
    colback=white,
    colframe=gray!50,
    boxrule=0.5pt,
    arc=2pt,
    left=6pt, right=6pt, top=4pt, bottom=4pt,
]
The value of \cat lies not in any single benchmark comparison, but in providing a \textbf{simple} recipe (using off-the-shelf mixers and existing infrastructure) to train models whose inference cost can be controlled at test time.
As it stands, \textbf{\textit{no} other sequence mixer provides such controllability natively}, and \cat \textit{wraps} around them as a \textit{meta}-sequence mixer to provide this ability.
\end{tcolorbox}

\textbf{Limitations and Future Work.}
Future work should explore how to provide a better trade-off despite a growing memory.
Another interesting direction is data-dependent adaptivity. \cat, as it stands, requires users to choose a chunk size appropriate for their compute and memory budgets.
Instead, one could post-train \cat to learn to allocate budget itself based on the context and task.
Such post-training would enable truly adaptive efficiency.
Finally, scaling \cat to larger model sizes and trillions of tokens, and evaluating on reasoning-heavy benchmarks (\texttt{GSM8K}, \texttt{MATH}, \texttt{GPQA}, \texttt{SWE-bench}), is beyond our limited compute budget and is left to future work
; we expect the benefits shown here to transfer at scale, as observed in prior work \citep{yang2025gated, dao2024transformers}.

\paragraph{Acknowledgments}
We would like to thank Neelabh Madan, Saksham Rastogi (\texttt{pogs}), Aastha Jain, Raghav Singhal, Zhixuan Lin, Mark Goldstein, Ethan Barron, Anirudh Buvanesh, Atharv Sonwane, Daman Arora, Divyam Madaan, Anshuk Uppal, William Merrill and Michael Hu for super useful discussions and feedback.
This work was partly supported by the NIH/NHLBI Award R01HL148248, NSF Award 1922658 NRT-HDR: FUTURE Foundations, Translation, and Responsibility for Data Science, NSF CAREER Award 2145542, NSF Award 2404476, ONR N00014-23-1-2634, Optum, and Apple.
We would also like to thank the support by IITP with a grant funded by the MSIT of the Republic of Korea in connection with the Global AI Frontier Lab International Collaborative Research.

%% file: sections/06_appendix.tex
\newpage
\appendix

\onecolumn

\newpage

\input{sections/07_extended_related_work}

\newpage
\section{More experiments}
\label{app:more_experiments}

\subsection{Long-context modeling and understanding}

\begin{table}[h]
\centering
\caption{LongPPL \cite{fang2025what} (we report \texttt{log\_loss}) and zero-shot evaluation on a suite of tasks from LongBench \cite{bai2023longbench} (upto 4K tokens) and common LM evals \cite{eval-harness}.
Refer to App. \Cref{tab:longbench} and \Cref{tab:lm_eval} for task-wise results, and to \Cref{app:datasets} for details.
All \cats are a single model, and perform competitively.
}
\footnotesize
\captionsetup{font=footnotesize,labelfont=footnotesize}
\setlength{\tabcolsep}{3pt}
\begin{tabular}{l|cc|c|c}
\toprule
& \multicolumn{2}{c|}{\textbf{LongPPL} $\downarrow$} 
& \textbf{LongBench} $\uparrow$
& \textbf{LM Evals}\footnotemark[\value{footnote}] $\uparrow$ \\
\cmidrule(lr){2-3} \cmidrule(lr){4-4} \cmidrule(lr){5-5}
\textbf{Model} & GovReport & PG19 & Avg. & Avg. \\
\midrule
\textit{Strict baselines}                        &  &  &  &  \\
Mamba2                       & 4.71 & 5.21 & 8.0  & 43.7 \\
\textsc{gdn}                 & 4.59 & 5.02 & 8.9  & 43.5 \\
\blue{\textsc{gdn}-2$\times$}& 4.23 & 4.86 & 8.1  & 43.8 \\
\blue{\textsc{gdn}-2$\times$ 2D}& 3.90 & 4.41 & 9.6  & 46.4 \\
\midrule
\cat-4                       & \textbf{2.96} & \textbf{4.20} & \textbf{13.9} & 43.9 \\
\cat-8                       & \ud{3.19} & \ud{4.30} & 12.1 & 44.1 \\
\cat-16                      & 3.66 & 4.57 & 9.5  & 44.3 \\
\cat-32                      & 4.36 & 4.92 & 7.9  & 45.1 \\
\midrule
\textit{Complementary}                        &  &  &  &  \\
Dense                        & 4.50 & 5.54 & 9.3  & 42.1 \\
Sparse-8                     & 5.09 & 5.54 & 9.3  & 43.2 \\
\textsc{gdn}-H 1:1           & 4.55 & 5.33 & 9.0  & 43.0 \\
\blue{\textsc{gdn}-H 1:4}    & 4.49 & 5.04 & 11.6 & 42.9 \\
\blue{\textsc{gdn}-H 1:5 G}  & 4.31 & 4.94 & \ud{12.2} & 43.7 \\
\blue{\textsc{gdn}-H 1:5 G 2D}& 4.02 & 5.04 & 11.6 & 45.8 \\
\bottomrule
\end{tabular}
\label{tab:longppl_longbench}
\end{table}
\footnotetext[\value{footnote}]{This evaluation only considers \textbf{short sequences} ($\leq 30$ tokens on average), which is \textit{not} indicative of long-context performance, as reported multiple times in the literature \cite{fang2025what, bertsch2025cracks}. Our claims concern long-context recall and modeling.}

\subsection{LongBench}
\label{app:long_bench}

\begin{table}[h]
\vspace{-0.6\baselineskip} %
\captionsetup{justification=RaggedRight,singlelinecheck=false,font=footnotesize,labelfont=footnotesize}
\raggedright
\setlength{\tabcolsep}{2pt}
\scriptsize
\centering
\setlength{\tabcolsep}{3pt} %
\begin{tabular}{l|cc|cc|cc|l}
\toprule
& \multicolumn{2}{c}{\textbf{Single-doc QA}} 
& \multicolumn{2}{c}{\textbf{Multi-doc QA}} 
& \multicolumn{2}{c}{\textbf{Few Shot}} 
& \textbf{Avg.} \\
\cmidrule(lr){2-3} \cmidrule(lr){4-5} \cmidrule(lr){6-7}
\textbf{Model} & \texttt{QAS} & \texttt{MQA} & \texttt{HQA} & \texttt{2WMQ} & \texttt{TQA} & \texttt{TREC} &  \\
\midrule
Dense                       & 3.9 & 12.2 & 6.9 & \textbf{10.8} & 11.2 & 10.6 & 9.3 \\
Sparse-8                      & 5.1 & 11.0 & 7.0 & \ud{10.6} & 10.5 & 5.6 & 9.3 \\
Mamba2                      & 4.1 & 11.9 & \textbf{7.6} & 7.6  & 9.0  & 7.6  & 8.0 \\
\textsc{gdn}               & \textbf{8.3} & \textbf{15.5} & 6.0 & 7.9  & 7.4  & 8.3  & 8.9 \\
\blue{\textsc{gdn-2$\times$}} & 4.1 & 11.8 & 6.7 & 9.6 & 9.8 & 6.8 & 8.1 \\
\blue{\textsc{gdn-2$\times$} 2D} & 4.6 & 14.0 & 6.9 & 8.9 & 11.3 & 12.1 & 9.6 \\
\textsc{gdn}-H 1:1           & 4.2 & 13.3 & 6.6 & 11.6 & 11.8 & 6.5  & 9.0 \\
\blue{\textsc{gdn}-H 1:4} & 4.6 & 13.0 & 7.0 & 10.4 & 10.6 & 24.2 & 11.6 \\
\blue{\textsc{gdn}-H 1:5 G} & 4.7 & 12.5 & 6.4 & 12.2 & 9.2 & 28.3 & \ud{12.2} \\
\blue{\textsc{gdn}-H 1:5 G 2D} & 3.9 & 12.9 & 8.3 & 9.1 & 11.9 & 23.7 & 11.6 \\
\midrule
\cat-4                      & \ud{5.6} & 12.7 & \ud{7.4} & 9.9  & \ud{12.1} & \textbf{35.6} & \textbf{13.9} \\
\cat-8                      & 5.5 & 11.0 & 6.1 & 8.0  & \textbf{12.4} & \ud{29.5} & 12.1 \\
\cat-16                     & 4.3 & \ud{14.1} & 6.1 & 5.6  & 10.5 & 16.6 & 9.5 \\
\cat-32                     & 4.7 & 11.0 & 7.0 & 6.6  & 10.0 & 8.3  & 7.9 \\
\bottomrule
\end{tabular}
\caption{Zero-shot evaluation of baselines on suite of tasks from LongBench \cite{bai2023longbench} up to $4$K tokens. Refer to \Cref{app:datasets}. \cat-4/8/16/32 are a single model.}
\label{tab:longbench}
\end{table}

\subsection{Recall}

\begin{table}[t]
\centering
        \caption{
        \blue{Zero-shot performance on real-world in-context recall tasks measured upto $4$K sequence lengths. H and G stand for Hybrid and Global respectively. \textbf{Fig. \ref{fig:pareto_frontier} gives an inference costs matched comparison using throughput.} All \cats here are a single model, whose costs can be changed at \textit{test-time} depending on downstream use-case.}
        }
        \setlength{\tabcolsep}{3pt}
        \begin{tabular}{lccc}
        \toprule
        \textbf{Model} & \textbf{SWDE} & \textbf{FDA} & \textbf{Avg.} $\uparrow$ \\
        \midrule
        \multicolumn{4}{l}{\textit{Strict baselines}} \\
        Mamba2    & 13.5 & 4.5 & 9.0 \\
        \textsc{gdn}       & 18.0 & 6.8 & 12.0 \\
        \blue{\textsc{gdn}-2$\times$}       & 24.0 & 11.0 & 17.5 \\
        \blue{\textsc{gdn}-2$\times$ 2D}       & 29.7 & 16.2 & 22.9 \\
        \midrule
        \cat-4   & \textbf{49.1} & \textbf{45.1} & \textbf{47.1} \\
        \cat-8   & 38.2 & 34.8 & 36.5 \\
        \cat-16  & 27.5 & 15.4 & 21.5 \\
        \cat-32  & 13.2 & 3.2  & 8.2  \\
        \midrule
        \multicolumn{4}{l}{\textit{Complementary }} \\
        Dense     & 43.4 & 19.7 & 32.0 \\
        \blue{Dense D/2}     & 32.0 & 22.0 & 27.0 \\
        \blue{Sparse-4}    & 36.0 & 16.0 & 26.0 \\
        Sparse-8    & 20.9 & 6.0 & 13.0 \\
        \textsc{gdn}-H 1:1      & 44.0 & 17.8 & 31.0 \\
        \blue{\textsc{gdn}-H 1:1 D/2}      & 17.0 & 3.2 & 10.0 \\
        \blue{\textsc{gdn}-H 1:4}      & 33.0 & 20.0 & 27.0 \\
        \blue{\textsc{gdn}-H 1:5 G}      & 38.0 & 25.0 & 32.0 \\
        \blue{\textsc{gdn}-H 1:5 G 2D}      & \ud{46.0} & \ud{36.0} & \ud{41.0} \\
        \bottomrule
        \end{tabular}
        \label{tab:swde_fda_results}
\end{table}

\subsection{RULER}

\begin{table}[h]
\footnotesize
\centering
\captionsetup{font=footnotesize} %
\caption{Accuracy on RULER \cite{hsieh2024ruler} S-NIAH-N benchmark. All \cats are a single model.}
\label{tab:s_niah}
\begin{tabular}{lccc}
\toprule
 & \multicolumn{3}{c}{\textbf{S-NIAH-N} ($\uparrow$)} \\
\cmidrule(lr){2-4}
\textbf{Model} & \textbf{1K} & \textbf{2K} & \textbf{4K} \\
\midrule
\textit{Strict baselines}           &          &          &          \\
Mamba2          & {97.7}       & 81.1         & 18.6         \\
\textsc{gdn}    & 84.7         & 69.1         & 13.6         \\
\blue{\textsc{gdn}-2$\times$}    & 78.0         & 61.4         & 29.0         \\
\blue{\textsc{gdn}-2$\times$ 2D}    & 99.3         & 97.0         & 72.0         \\
\midrule
\cat-4          & 96.0         & \ud{97.0} & \textbf{96.0} \\
\cat-8          & 90.0         & {93.0}     & \ud{91.0}     \\
\cat-16         & 76.0         & 72.0          & 70.0          \\
\cat-32         & 60.0         & 37.0          & 31.0          \\
\midrule
\textit{Complementary}           &          &          &          \\
Dense           & 96.0         & 92.0         & 43.0         \\
Sparse-8          & 51.2         & 46.2         & 5.0          \\
\textsc{gdn}-H 1:1 & \ud{99.0} & 97.0 & 44.0     \\
\blue{\textsc{gdn}-H 1:4}    & 98.0         & 96.0         & 35.8         \\
\blue{\textsc{gdn}-H 1:5 G}    &  98.3     & 93.0         & 23.2         \\
\blue{\textsc{gdn}-H 1:5 G 2D}    & \textbf{99.5}         & \textbf{99.3}         & 43.8         \\
\bottomrule
\end{tabular}
\end{table}

\newpage
\subsection{Short-context language understanding evaluations}

\begin{table}[h]
\scriptsize
\centering
\setlength{\tabcolsep}{3pt} %
\begin{tabular}{l|cccccc|c}
\toprule
\textbf{Model} & \textbf{HS}$\uparrow$ & \textbf{PQ}$\uparrow$ & \textbf{AE}$\uparrow$ & \textbf{AC}$\uparrow$ & \textbf{WG}$\uparrow$ & \textbf{OQA}$\uparrow$ & \textbf{Avg.}$\uparrow$ \\
\midrule
Dense & 34.8 & 65.6 & 56.7 & 24.4 & 51.1 & 20.0 & 42.1 \\
Sparse-8 & 35.6 & 66.8 & 57.3 & 25.4 & 51.1 & 22.8 & 43.2 \\
Mamba2 & 36.1 & 67.0 & 59.2 & 26.5 & 51.9 & 21.6 & 43.7 \\
\textsc{gdn} & 36.1 & 66.8 & 58.7 & 25.2 & 51.6 & 22.8 & 43.5 \\
\blue{\textsc{gdn}-2$\times$} & 35.9 & 67.4 & 58.6 & 27.2 & 51.8 & 21.8 & 43.8 \\
\blue{\textsc{gdn}-2$\times$ 2D} & 37.5 & 69.9 & 63.6 & 29.4 & 52.8 & 25.6 & 46.5 \\
\textsc{gdn}-H 1:1 & 36.8 & 66.3 & 56.4 & 25.8 & 52.1 & 20.4 & 43.0 \\
\blue{\textsc{gdn}-H 1:4} & 34.8 & 67.0 & 57.0 & 26.5 & 50.3 & 22.0 & 42.9 \\
\blue{\textsc{gdn}-H 1:5 G} & 36.0 & 67.6 & 57.4 & 26.6 & 51.5 & 23.4 & 43.7 \\
\blue{\textsc{gdn}-H 1:5 G 2D} & 38.6 & 70.1 & 62.3 & 27.8 & 52.7 & 23.6 & 45.8 \\
\midrule
\cat-4 & 35.6 & 66.4 & 59.5 & {27.1} & 51.5 & 23.4 & 43.9 \\
\cat-8 & 35.4 & 66.8 & 60.1 & {27.4} & 51.3 & 23.6 & 44.1 \\
\cat-16 & 35.5 & {67.3} & {60.2} & 27.0 & {52.0} & {23.8} & {44.3} \\
\cat-32 & 35.9 & {68.2} & {61.0} & 27.0 & {53.6} & {25.0} & {45.1} \\
\bottomrule
\end{tabular}
\captionsetup{font=footnotesize,labelfont=footnotesize}
\caption{Zero-shot accuracy on common-sense reasoning benchmarks.
However, note that these evaluations considers short sequences only ($\leq30$ tokens on average).
Hence, we test language understanding on longer contexts in \Cref{tab:longppl_longbench} on LongBench \citep{bai2023longbench}, LongPPL \cite{fang2025what} and test in-context recall on real world tasks \citep{arora2023language} in the main text.
\Cref{app:datasets} expands the acronyms in \cref{tab:lm_eval}.
}
\label{tab:lm_eval}
\end{table}

\newpage
\subsection{Parameter-Matched Results}
\label{app:parameter_matched}

Here, we report results for parameter-matched comparisons to complement our inference-cost-based analysis in the main paper. 
We acknowledge that both parameter count and inference cost are axes for model comparison.
That being said, quality \textit{at a given inference-cost} is the correct comparison criterion in the current age, where models spend more time doing inference as chat-assistants, coding agents and more -- and what the end-user eventually cares about is quality for cost (parameter count matters in so far as they affect the inference cost).

\begin{figure}[H]
    \centering
    \includegraphics[width=0.5\linewidth]{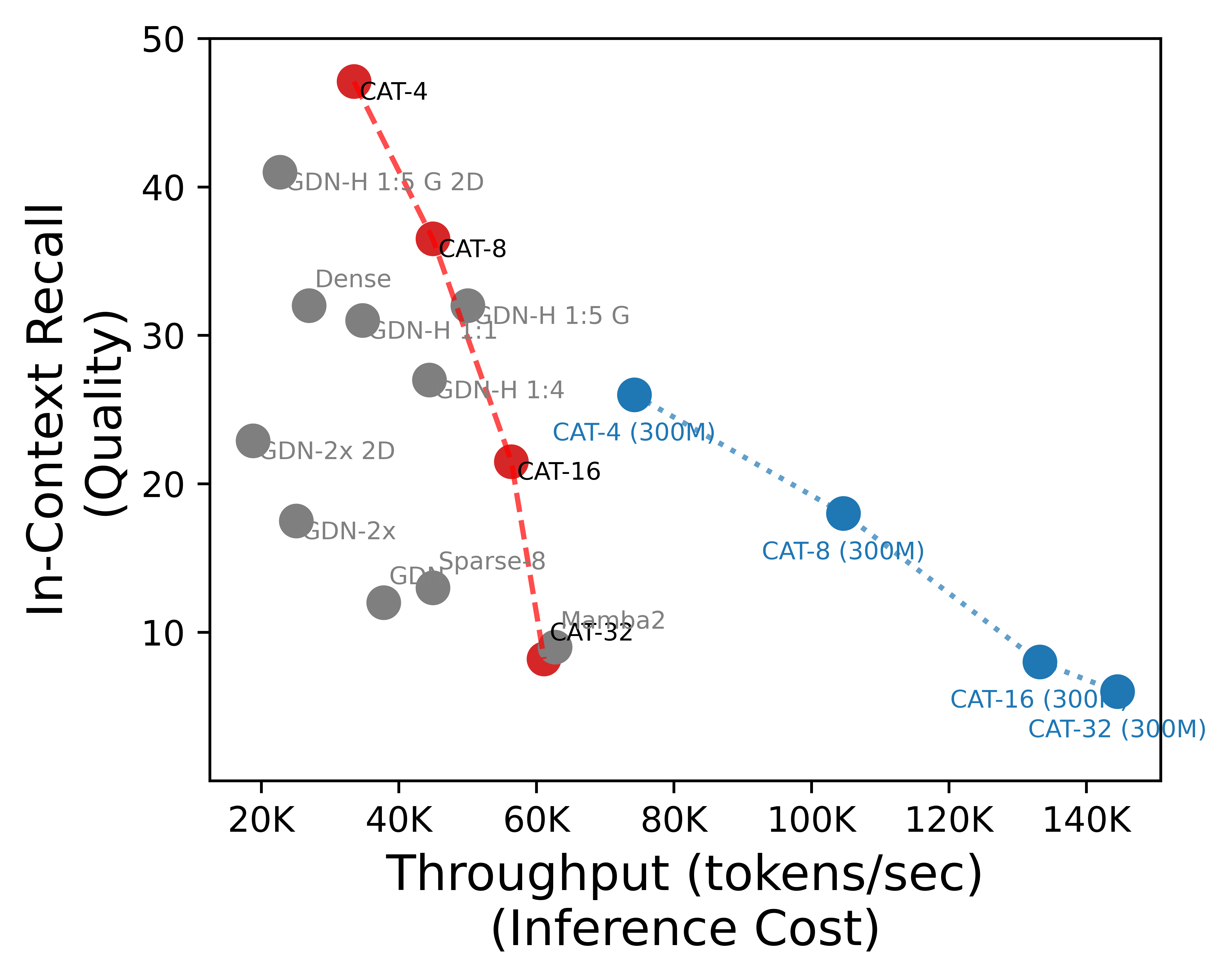}
    \caption{We plot smaller-parameter \cat models (\textbf{\textcolor{blue}{blue}} dots) alongside the models in \Cref{fig:pareto_frontier}. We note smaller-parameter \cat models have the least inference-costs (highest throughput) among all models considered in our evaluation.
    Because the comparison is quality \emph{at a given throughput}, comparing these \textbf{\textcolor{blue}{blue}} dots to the other models is comparing apples-to-oranges: smaller-parameter \cats operate at a substantially lower inference costs.
    Further, it showcases the point about improved trade-offs (\textbf{\textcolor{blue}{blue}} $\rightarrow$ \textbf{\textcolor{red}{red}}) when increasing parameters in \cat: the quality improves without taking a proportional hit to the throughput (or inference cost). 
}
    \label{fig:parameter_matched_figure}
\end{figure}

\begin{figure}[H]
    \centering
    \includegraphics[width=0.9\textwidth]{figures/main_results_parameter_matched.png}
    \caption{Lower parameter \cats comparison.}
    \label{fig:parameter_matched_figure_main_results}
\end{figure}

\newpage

\begin{figure}[h]
    \centering
    \begin{minipage}[t]{0.5\linewidth}
        \centering
        \includegraphics[width=\linewidth]{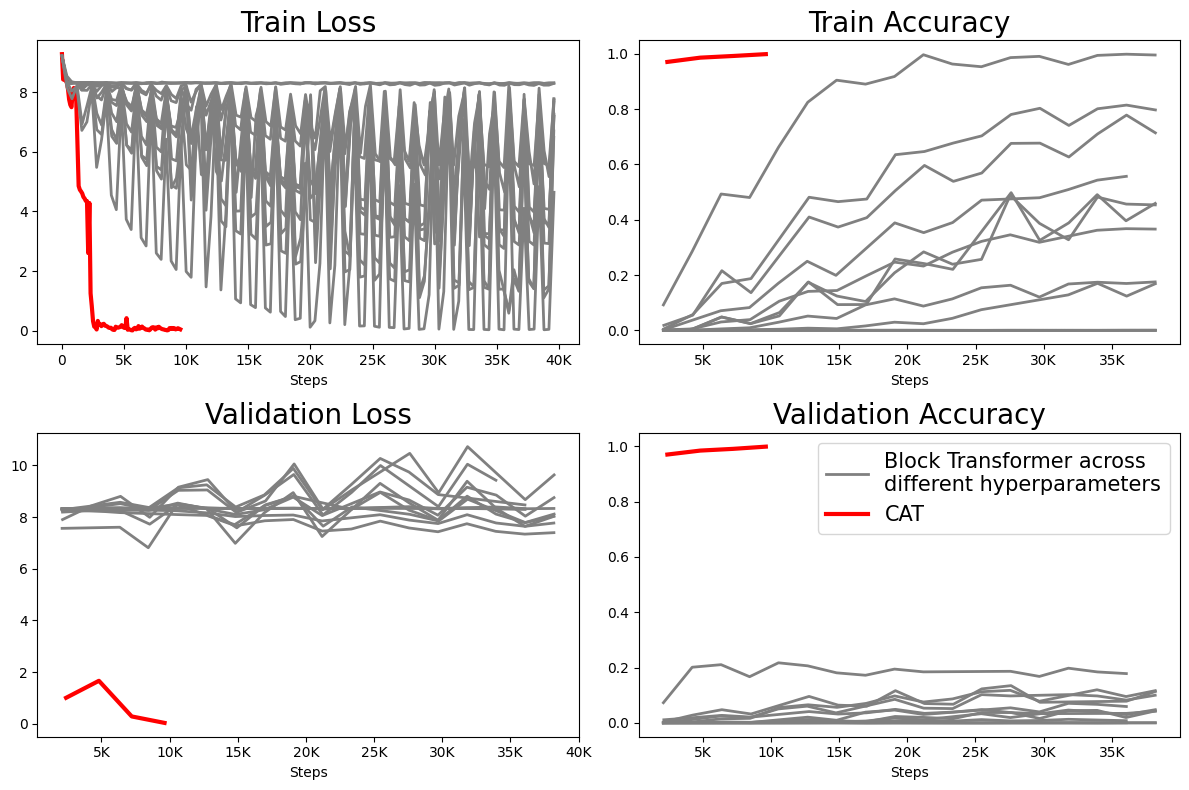}
        \caption{
        Block Transformer \cite{ho2024block, yu2023megabyte} (across different configurations and hyperparameters) fails to solve a simple MQAR task with only 4 key-value pairs tested on modest sequence length of 256 tokens.
        Note that training of \cat stops when it solves the task perfectly.
        }
        \label{fig:block_transformer_fails}
    \end{minipage}
    \hfill
    \begin{minipage}[t]{0.45\linewidth}
        \centering
        \includegraphics[width=\linewidth]{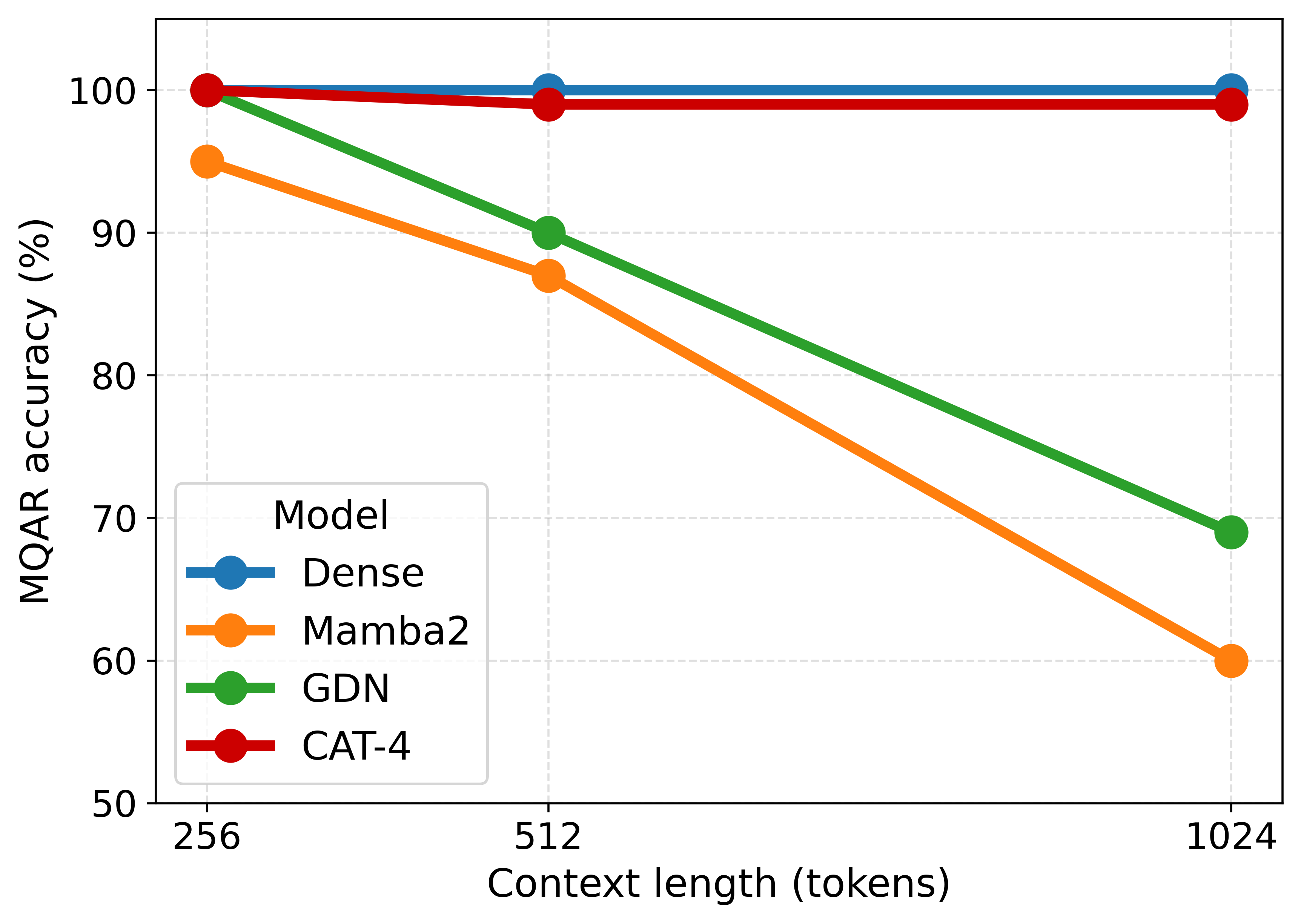}
        \caption{Comparison of different architectures across sequence lengths on MQAR task. 
        We measure test-accuracy on the hardest subset.
        All architectures are memory matched in bytes at every point (except dense transformer).}
        \label{fig:mqar_cat_gdn_mamba}
    \end{minipage}
\end{figure}

\subsection{Synthetic tasks}
\label{app:synth_task}

We begin by evaluating models on the multi-query associative recall (MQAR) task \cite{arora2023zoology}, a standard diagnostic benchmark for testing in-context recall. 
Our setup
mostly follows \cite{arora2024simple}: except we train and evaluate models on sequences $4\times$ the standard sequence length, upto 1024 tokens containing maximum amount of key-value pairs possible.
We do not test length generalization here.
We compare \cat with dense attention and linear methods, namely Mamba2 \cite{dao2024transformers} and Gated DeltaNet \cite{yang2025gated}. 
Crucially, we ensure all models get equal memory (or state size) down to the level of bytes at each sequence length (except dense) to test recall-memory trade-offs. 
For linear models, this required scaling up state sizes to match \cat at every sequence length.
All models were grid-searched for best hyper-parameters.
We specifically only compare with linear sequence mixers having length independent costs since they are the only strict baselines to \cat (we discuss this in next section).
\cat maintains strong recall performance across all sequence lengths. 
We especially note the following:

\begin{tcolorbox}[
    colback=gray!5,
    colframe=gray!50,
    boxrule=0.5pt,
    arc=2pt,
    left=6pt, right=6pt, top=4pt, bottom=4pt,
    fontupper=\footnotesize
]
Sequence mixers like dense attention or linear recurrences take an \textit{extreme} route for long-context modeling: either the memory keeps growing \textit{aggressively}, or the memory is fixed.
\cats take a \textbf{middle-ground}, where the memory increases \textit{gracefully} by constant factor less than dense.
This flexible memory learns long-context correlations better, resulting in better recall despite taking the same memory (better trade-off) compared to existing fixed-memory sequence mixers.
\end{tcolorbox}

To rule out any memory discrepancy, (Fig.~\ref{fig:generation_throughput}) evaluates on MQAR task \citep{arora2023zoology}, matching memory budgets down to the level of bytes, and stress-tests models up to $1$K sequence length ($5\times$ standard); \Cref{fig:mqar_cat_gdn_mamba} in reports results.
Baselines are grid-searched over learning rates.
Linear models collapse at longer contexts, while \cats remain near-perfect, thanks to the flexible yet efficient memory scaling.
We use the same setup in App. \ref{app:sparse_fails_mqar}.

\subsection{Comparison with MegaByte/Block Transformer}
\label{app:block_transformer_fails}

The MegaByte/Block Transformer \citep{ho2024block, yu2023megabyte} has elements similar to \cat but fail to solve a simple in-context recall task in \cref{fig:block_transformer_fails} across different hyperparameters and architecture configurations due to the fixed memory bottleneck.
In fact, the block transformer overfits on the task.
\cats alleviate the memory bottleneck with a gracefully growing memory, allowing it to solve the task, with even lower memory requirements.

In figure \ref{fig:block_transformer_fails}, we evaluate in-context recall ability for Block Transformer architectures \cite{ho2024block, yu2023megabyte}, that model chunks of tokens similar to \cats but with a subtle but salient difference in the architecture circuit (that we explain below).
For this experiment, we test on the MQAR task (a synthetic needle-in-haystack task \cite{arora2023zoology}) on a modest sequence length of 256. 
We test the accuracy of retrieving just 4 needles. 
We parametrize components of Block Transformer that is: global model and local model using a transformer, the embedder is a look-up table or a transformer. 
We keep the patch size/chunk size as 4 – same as \cat. 
We keep the identical training setup for both architectures.
We grid search for hyper-parameters (\texttt{lr}, \texttt{hidden\_size}, and embedder parameterization), even \textbf{using more memory} than the \cat baseline, in its global decoder.
Even in these simple settings and added advantage, Block Transformer \cite{ho2024block, yu2023megabyte} fails to solve the task (fig. \ref{fig:block_transformer_fails}) – instead the model starts to memorize the train points, as seen from train loss and train accuracy -- train metrics keep getting better, however, test metrics suffer. 

\Cats directly pass all the “local” patch/chunk representations directly to the decoder, unlike the block transformer that forces the history to be compressed into fixed dimensional representation.
This design choice helps \cat \textit{alleviate the memory bottleneck} that \cite{ho2024block} suffers from where the architecture must compress everything from the past into a single "global" representation to generate the next chunk. 
Note that this different design choice in \cats does not introduce any memory/compute overhead compared to Block Transformer \cite{ho2024block}, it just changes the circuit of the architecture. In fact, \cats don’t utilize three different components (embedder, global decoder, local decoder) – it only uses a compressor and a decoder, reducing the design space and (significant) parameter requirements further.

\newpage
\subsection{RULER benchmark}
\label{app:niah}

\Cref{tab:s_niah} (in the main text) reported results on RULER \cite{hsieh2024ruler} single-needle tasks: S-NIAH-N (recall number from the context).
We observed linear recurrent models (Mamba2, \textsc{gdn}) struggle at longer contexts, and while \textsc{gdn}-Hybrid narrows the gap with dense transformers, performance still drops at longer contexts.
\Cats-4/8/16 outperform the efficient baselines as context length increases, showing slower degradation with length, even compared to the dense transformer.
This slow degradation can possibly be attributed to reduced sequence length in \cat that leads to fewer \textit{distractions} for attention \citep{barbero2024transformers, vasylenko2025long, chiang2022overcoming, golovneva2025multi}.
Further, large-chunk \cat underperforms at short contexts but interestingly surpass baselines at long ones.
One reason why large-chunk \cat underperforms could be due to ineffective compression -- due to larger chunks, the compressor in \cat is not always able to surface the \textit{right} information in the chunk representation for accurate retrieval. 
More pre-training or finetuning on specific task data alleviates this problem for large chunk \cat (see App. \ref{app:finetuning_cats}).
That being said, there is an upper limit to how much information fixed sized chunk representations can practically learn to hold for large token chunks.

\Cref{tab:babilong} further reports results on the harder S-NIAH-U (recall a long alpha-numeric string or UUID).

\begin{table}[h]
\centering
\caption{Accuracy on RULER \cite{hsieh2024ruler} S-NIAH-U benchmark.}
\label{tab:babilong}
\begin{tabular}{lccc}
\toprule
 & \multicolumn{3}{c}{\textbf{S-NIAH-U}} \\
\cmidrule(lr){2-4}
\textbf{Model} & \textbf{1K} & \textbf{2K} & \textbf{4K} \\
\midrule
Dense           & \textbf{93.6} & 55.7 & 19.8 \\
Sparse          & 12.8          & 1.4  & 0.8  \\
Mamba2          & 46.7          & 4.6  & 1.0  \\
\textsc{gdn}    & 38.9          & 2.6  & 2.0  \\
\textsc{gdn}-H 1:1 & 50.9      & 5.6  & 2.6  \\
\midrule
\cat-4          & \ud{79.6}     & \textbf{59.3} & \ud{46.5} \\
\cat-8          & 68.1          & \ud{57.5}     & \textbf{47.3} \\
\cat-16         & 10.0          & 6.6           & 3.8           \\
\cat-32         & 0.0           & 0.0           & 0.0           \\
\bottomrule
\end{tabular}
\end{table}

\newpage
\subsection{Recall evaluation}
\label{app:recall_evaluation}

Here, we evaluate all baselines on all datasets from the EVAPORATE suite of tasks that tests for real-world in-context recall.

\begin{table}[h]
\centering
\begin{tabular}{lcccccc}
\toprule
\textbf{Model} & \textbf{SWDE} & \textbf{FDA} & \textbf{Squad} & \textbf{TriviaQA} & \textbf{Drop} & \textbf{Avg.} \\
\midrule
Dense     & 43.4 & 19.7 & \ud{31.0} & 15.0 & \ud{19.4} & \ud{26.7} \\
Sparse    & 20.9 & 6.0  & 20.7 & 15.2 & 19.3 & 16.4 \\
Mamba2    & 13.5 & 4.5  & 24.9 & 13.9 & 17.8 & 14.9 \\
\textsc{gdn}       & 18.0 & 6.8  & 25.5 & \textbf{15.5} & 17.2 & 16.6 \\
\textsc{gdn-h 1:1}    & 44.0 & 17.8 & \textbf{32.9} & \ud{15.4} & \textbf{19.8} & 26.0 \\
\midrule
\cat-4    & \textbf{49.1} & \textbf{45.1} & 28.3 & 15.0 & 17.9 & \textbf{31.1 }\\
\cat-8    & \ud{38.2} & \ud{34.8} & 25.9 & 14.0 & 18.3 & 26.2 \\
\cat-16   & 27.5 & 15.4 & 20.4 & 14.8 & 16.9 & 18.9 \\
\cat-32   & 13.2 & 3.2  & 15.8 & 13.0 & 14.3 & 11.9 \\
\bottomrule
\end{tabular}
\caption{Zero-shot performance on real-world in-context recall tasks from EVAPORATE suite, measured upto $4$K sequence lengths.
Note that only SWDE and FDA have long token sequences among the datasets in the suite (others have an average length of $\leq300$ tokens \cite{arora2024just}). 
}
\label{tab:all_evaporate_results}
\end{table}

\newpage

\subsection{\cat as a layer}
\label{app:cat_as_layer}

While \cat presented in the paper is a separate meta-sequence mixer, one can take the core concepts and instantiate \cat as a layer that can be swapped in any sequence model as a drop-in replacement.
This can unlock lots of interesting possibilities starting with creating hybrid as well as adaptive architectures that mixes \cat layers alongside dense attention, or perhaps even linear attention.
We leave this open for future work.

\blue{To instantiate \cat as a seperate layer in itself, we parameterize the \textit{compressor} as a simple linear projection. 
We use the dense attention mechanism itself as the \textit{decoder}.
Before applying the compression and decoding from compressed chunk representations, we artificially up-project the input embeddings in the layer -- this is done following the observation in the main paper that decoding from compressed representations requires higher dimensionality.
We will release the implementation in our code.
\Cref{tab:cat_as_layer} reports MQAR accuracy when \cat is used as a layer.
We use a fixed chunk size of 4 in this experiment.
We use 2 layers of \cat.
Rest of the setup follows \cite{arora2024simple}.
}

\begin{table}[h]
\centering
\begin{tabular}{l|c|r}
\textbf{Method} & \textbf{Solves?} & \textbf{State Size} \\
\hline
Dense & \cmark & $16384$ \\
\cat & \cmark & $4096$ \\
\cat (layer) & \cmark & $4096$ \\
\end{tabular}
\caption{\blue{\cat instantiated as a seperate layer solves the MQAR task.}}
\label{tab:cat_as_layer}
\end{table}

\subsection{\cat is a meta sequence mixer}
\label{app:cat_as_meta_mixer}

\cat has two components: a compressor and a decoder -- each of these could make use of any sequence mixers, such as linear attention. Here we provide preliminary result on the MQAR task where the decoder in \cat is a \textsc{gdn-hybrid} architecture \citep{yang2025gated} (having a 1:1 dense-to-linear ratio). This new architecture solves this task, empirically demonstrating the use of a hybrid sequence mixer inside of \cat: meaning rather than \cat being a strict competitor to hybrids, \textbf{\cat is complementary}. Further, the use of a different sequence mixers inside of \cat can unlock the test-time control of efficiency with those sequence mixers (e.g., \textsc{GDN-Hybrid} in this case).

\blue{
\cat solves the MQAR task when the decoder is instantiated as a GDN-Hybrid architecture (1:1 dense-linear ratio) with 2 layers. We use the same setup described in \cite{arora2024simple}: using sequences upto 256 with maximum key-values in the sequence. The chunk size used is set to 4.
}

\begin{table}[h]
\centering
\begin{tabular}{l|c}
\textbf{Method} & \textbf{Solves?} \\
\hline
Dense & \cmark \\
\cat & \cmark \\
\cat (\textsc{gdn-hybrid} decoder) & \cmark \\
\end{tabular}
\caption{\blue{The decoder in \cat is replaced with a \textsc{gdn-hybrid} architecture. The resulting \cat architecture solves the MQAR task.}}
\label{tab:cat_as_layer}
\end{table}

\blue{Further, we train a \cat{} variant with \textsc{gdn}-1:1 as the decoder on 5B tokens of FineWeb at 1K context. 
\cat{} reduces memory cost when applied on top of \textsc{gdn-h 1:1}, and the resulting model even outperforms \cat{} with a dense attention decoder -- further evidence that \cat{} and efficient sequence mixers are complementary.}

\newpage

\newpage
\subsection{\cats scale as well as their dense counterparts}

\begin{figure}[h]
    \centering
    \includegraphics[width=0.5\linewidth]{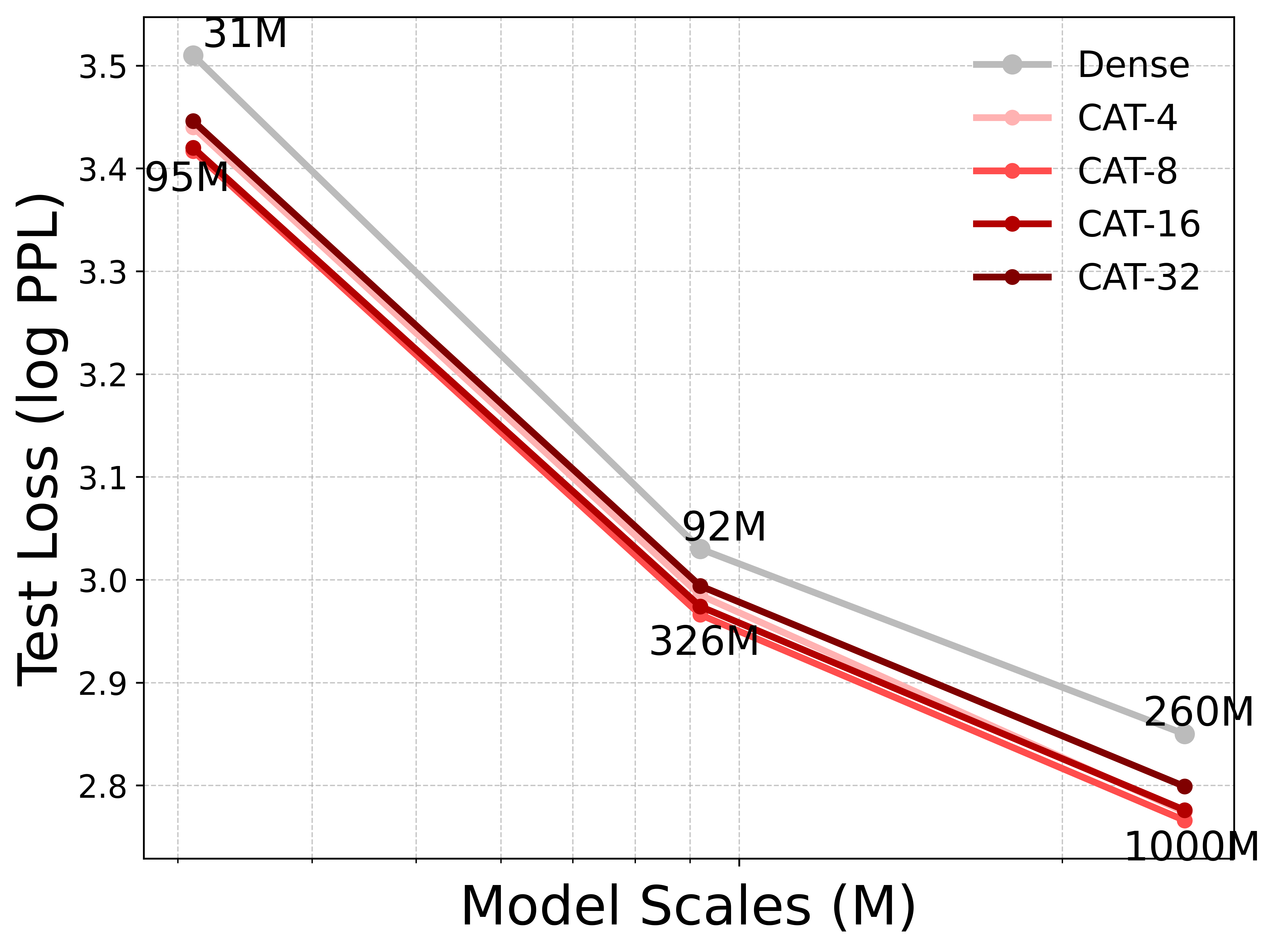}
    \caption{\cats scale like their dense transformer counterparts while being up to $3\times$ faster and $9\times$ more memory-efficient. All \cat curves come from a single model, evaluated at different chunk sizes. \textbf{Note that while \cat occupies more parameters, it is still both compute and memory efficient compared to the densee transformer at every scale.}}
    \label{fig:scaling_law}
    \vspace{-10pt} %
\end{figure}

\Cref{fig:scaling_law} demonstrates that \cats scale similar to their dense transformer equivalents.
We evaluate against three dense transformer scales $\{31M, 92M, 260M\}$, with their \cat equivalents containing parameters $\{95M, 326M, 1B\}$.
All models were trained for 15B tokens, under the setup in \cref{sec:experiments}.

\begin{blueblock}
\subsection{Across chunk analysis}
\label{app:across_chunk_anal}

We provide how the validation loss changes within a chunk in \Cref{fig:chunk_anal}. 
We provide averaged results across all chunks. We provide different curves for each chunk size.

Interestingly, across all chunk sizes, the loss is highest when decoding the first token from the compressed representations only. 
After that token is decoded, the loss decreases steadily as \cat keeps decoding tokens from both compressed representation and raw tokens that appear before inside the chunk.

\begin{figure}[h]
    \centering
    \includegraphics[width=0.5\linewidth]{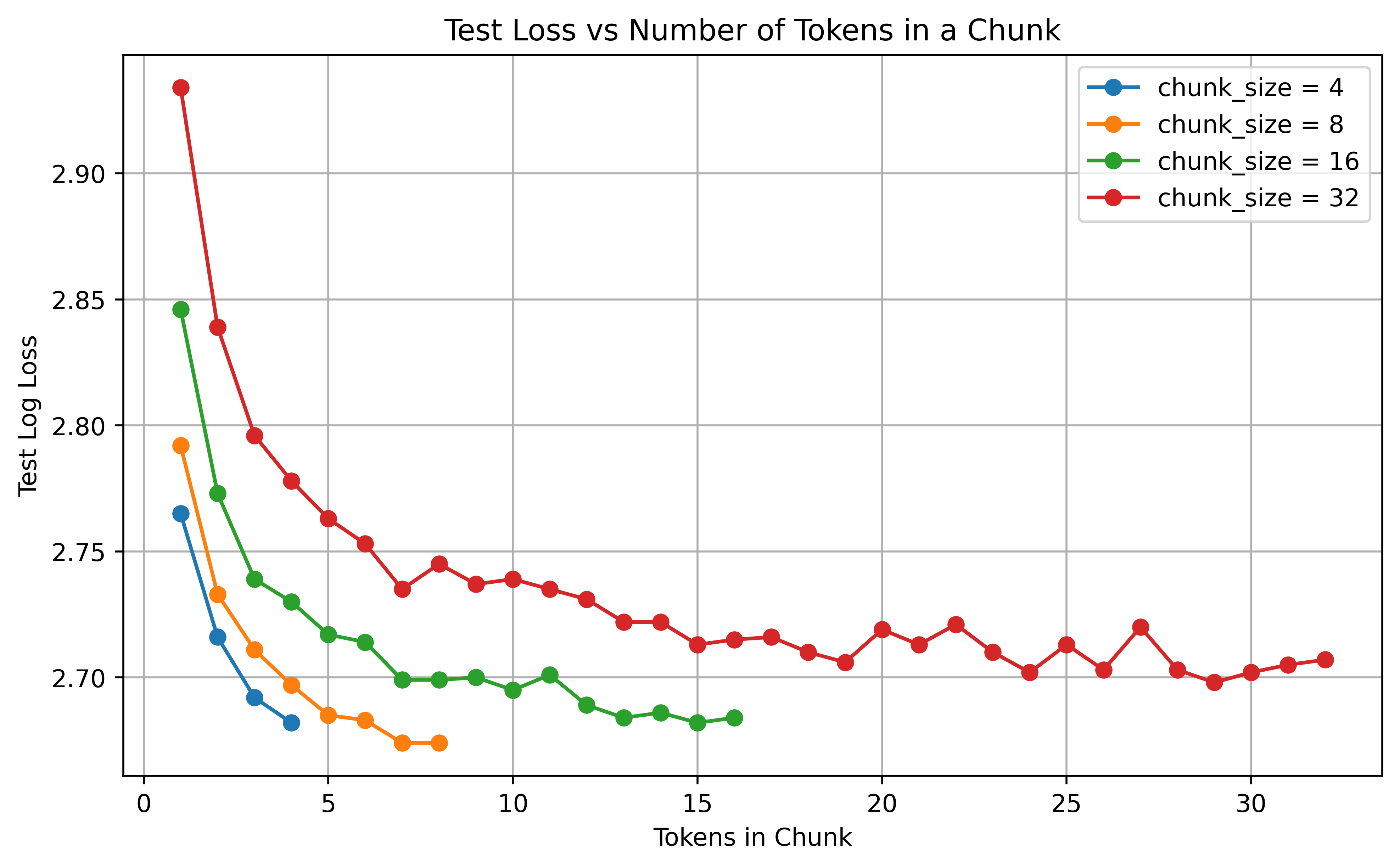}
    \caption{Chunk Analysis}
    \label{fig:chunk_anal}
\end{figure}

\end{blueblock}

\newpage

\subsection{Finetuning \cats on S-NIAH-U}
\label{app:finetuning_cats}

\begin{figure}[h]
    \centering
    \includegraphics[width=0.5\linewidth]{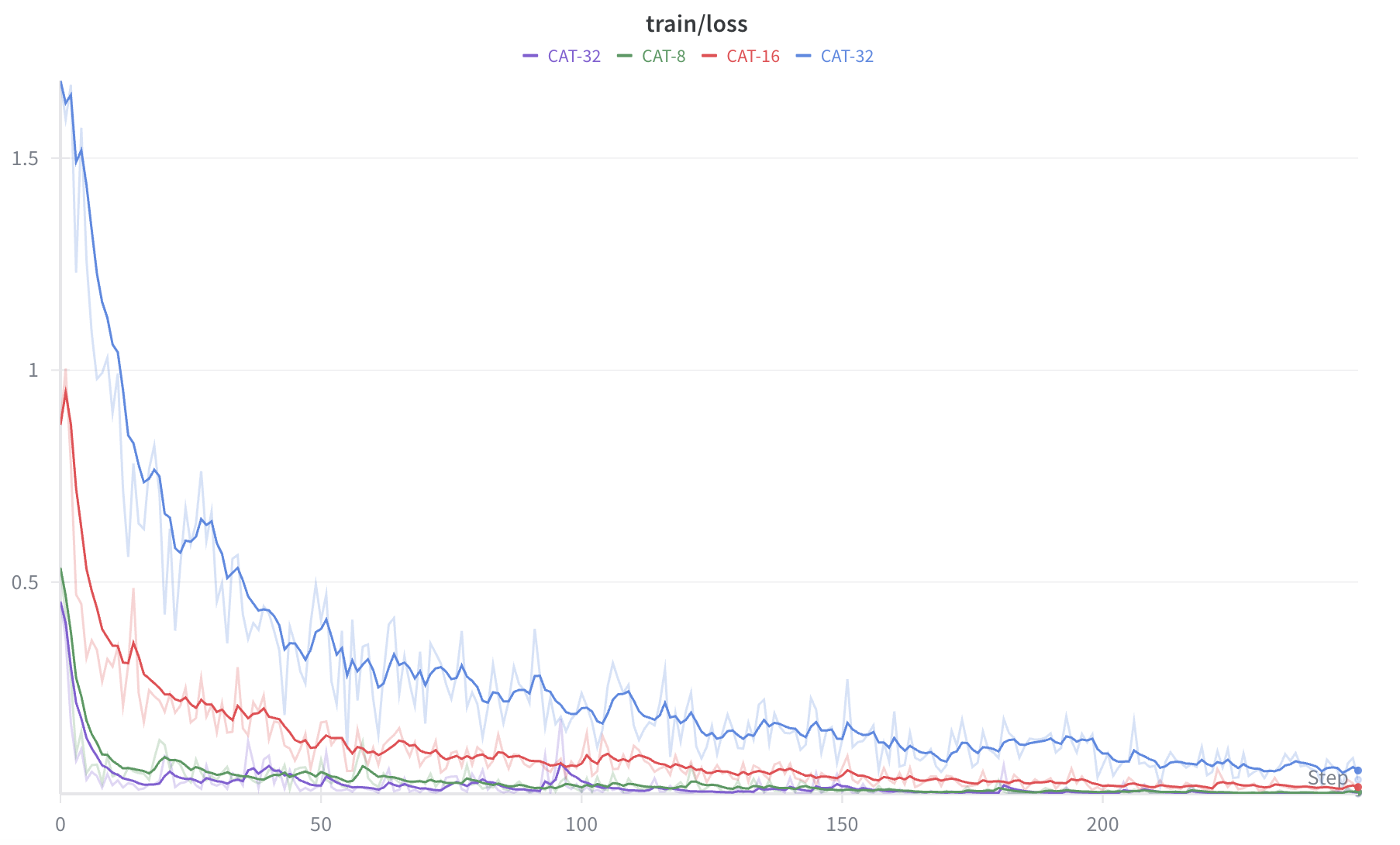}
    \caption{Loss curves when finetuning different \cats on samples from S-NIAH-U task.}
    \label{fig:finetuning_cats}
\end{figure}

S-NIAH-U is a task where model needs to recall 32 token long UUID strings from the long context.
This section reports performance of \cats after task specific finetuning on samples from S-NIAH-U.
We only apply the loss on tokens that appear in the answer span.
\Cref{tab:finetuning_cats} reports these results.
This is accompanied by loss curves for different \cats depending on chunk size in \Cref{fig:finetuning_cats} on this task.

We observe two things: (i) after finetuning, performance goes up significantly for all chunk sizes. This signifies as chunk size increased, compressor in \cats, before finetuning, was not surfacing the \textit{right} information in the chunk representation.
(ii) the loss curves during finetuning indicate the same as well, however it still does not go completely to zero, especially for \cat-32. This indicates that there are limits to what information a fixed sized chunk representation can practically learn to surface, justifying its sub-par accuracy on the task.

This problem of not surfacing the \textit{right} information in the chunk representation could be alleviated by more and longer pre-training, or choosing smaller chunk sizes for tasks that require accurate recall.

\begin{table}[h]
\scriptsize
\centering
\begin{tabular}{lcc}
\toprule
\textbf{Model} & \textbf{Before} & \textbf{After} \\
\midrule
\cat-4  & 46.5 & 97.1 \\
\cat-8  & 47.3 & 97.0 \\
\cat-16 & 3.8  & 94.2 \\
\cat-32 & 0.0  & 64.3 \\
\bottomrule
\end{tabular}
\caption{Performance on 4K sequence length before and after finetuning for different \cat variants.}
\label{tab:finetuning_cats}
\end{table}

\newpage
\subsection{Comparison with Training-Free Block-Sparse Attention}
\label{app:training_free}

An interesting comparison is with training-free block-sparse attention, where the past KV cache is chunked into blocks and top-$K$ block selection is applied at inference time \cite{jiang2024minference}. 
We emphasize that this approach is \emph{complementary} to \cat rather than a direct competitor: training-free block-sparse attention yields only \emph{compute} savings, while \cat is designed to deliver both \emph{compute and memory} savings during inference (see Table~\ref{tab:related_work} for a detailed comparison with closely related works). 
In long-context inference regimes on modern hardware, memory savings are typically the more critical bottleneck, since available FLOPs substantially outpace available GPU memory.

Nevertheless, for completeness, we evaluate training-free block-sparse attention on the SWDE recall task. 
We chunk the past KV cache into blocks and perform top-$K$ block selection \cite{jiang2024minference} with the reference implementation from the \texttt{nano-sparse-attention} repository~\cite{nawrot2024nanosparseattention}. 
We fix \texttt{block\_size}~$=32$ and sweep the top-$K$ parameter. We also include the trainable strided sparse attention models~\cite{child2019generating} from Table~\ref{tab:swde_fda_results} for reference.

For block-sparse attention, the per-token FLOPs reduction is $\mathcal{O}\!\left(N / (\texttt{block\_size} \cdot \texttt{top\_K})\right)$ with sequence length $N = 4096$ and $\texttt{block\_size} = 32$. For CAT, both the per-token FLOPs reduction and the memory reduction are $\mathcal{O}\!\left(N / \texttt{chunk\_size}\right)$.

\begin{table}[h]
\centering
\small
\setlength{\tabcolsep}{6pt}
\renewcommand{\arraystretch}{1.15}
\begin{tabular}{lccc}
\toprule
\textbf{Method} & \makecell{\textbf{Theoretical} \\ \textbf{Memory Red.}} & \makecell{\textbf{Theoretical} \\ \textbf{Per-Token FLOPs Red.}} & \textbf{SWDE} \\
\midrule
Dense & $1\times$ & $1\times$ & 43.4 \\
\midrule
\multicolumn{4}{l}{\emph{Block-Sparse Attention (Training-Free)}} \\
\quad top-32 & $1\times$  & $4\times$  & 43.0 \\
\quad top-16 & $1\times$  & $8\times$  & 34.4 \\
\quad top-8  & $1\times$  & $16\times$ & 18.9 \\
\quad top-4  & $1\times$  & $32\times$ & 8.0  \\
\midrule
\multicolumn{4}{l}{\emph{Trained Strided Sparse Attention}} \\
\quad Sparse-4 & $4\times$ & $4\times$ & 36.0 \\
\quad Sparse-8 & $8\times$ & $8\times$ & 20.9 \\
\midrule
\multicolumn{4}{l}{\emph{CAT (Ours)}} \\
\quad CAT-4   & $4\times$   & $4\times$   & \textbf{47.1} \\
\quad CAT-8   & $8\times$   & $8\times$   & \textbf{36.5} \\
\quad CAT-16  & $16\times$  & $16\times$  & \textbf{21.5} \\
\quad CAT-32  & $32\times$  & $32\times$  & \textbf{8.2}  \\
\bottomrule
\end{tabular}
\caption{Comparison of \cat against training-free block-sparse attention and trainable strided sparse attention on the SWDE recall task ($N = 4096$). Training-free block-sparse attention provides only FLOPs savings ($1\times$ memory reduction), whereas \cat achieves both compute and memory savings at a better recall–cost trade-off.}
\label{tab:block-sparse-comparison}
\end{table}

\paragraph{Takeaways.} Two observations follow from Table~\ref{tab:block-sparse-comparison}:
\begin{itemize}
    \item Training-free block-sparse attention provides FLOPs savings but \emph{no memory savings}, since the full KV cache must still be retained for top-$K$ block selection.
    \item \cat achieves \emph{both} compute and memory savings, and attains a strictly better recall–cost trade-off across all matched compression ratios, a benefit we attribute to end-to-end training of the compression and decoding, and thus not resulting in a train-test mismatch that happens with \textit{training-free} approaches.
\end{itemize}
This further supports our view that block-sparse attention is complementary to \cat rather than a competing baseline: assuming \cat's decoder uses a sequence mixer with length-dependent cost (as in our default dense attention decoder), block-sparse selection could in principle be layered on top of \cat to obtain additional FLOPs reductions on the already-compressed KV cache.

\newpage
\section{Implementation details and PyTorch style pseudo-code}
\label{app:pseudo_code}

In this section, we discuss some implementation details regarding \cats.
We repeat some text presented in the main paper to be self-contained below.

\subsection{Training}

\paragraph{Training:} While \cats are simple and build on dense transformer abstractions, their naive PyTorch training implementation is very inefficient.

Note that compression of chunks of tokens is efficient since it can be done in parallel, specifically using \texttt{torch.vmap}($f_\theta(\mbc_i)$) for all chunks $\mbc_i$. 
This costs a total of $O(\frac{N}{C}\cdot C^2)=O(NC)$ in self-attention compute, which is much better than $O(N^2)$.

But, computing logits for tokens in chunk $\mbc_i$, that is computing $g_\theta (\mbc_i \mid f_\theta(\mbc_1) \cdots f_\theta(\mbc_{i-1}))$ can be non-trivial since for chunk $\mbc_i$, we have $i-1$ past chunk representations $\{f_\theta(\mbc_1),f_\theta(\mbc_2)\dots f_\theta(\mbc_{i-1})\}$. 
In other words, there are different number of past chunk representations for every chunk, making shapes variable and as a result, harder to parallelize computation of logits.
One could employ a python loop and compute logits for every chunk sequentially, but that would be slow and won't scale. 
In fact, even if one manages to compute logits for every chunk in parallel, the total self-attention operations in the decoder would be $O(\sum_{i=1}^{\frac{N}{C}}(i+C)^2)=O((\frac{N}{C})^3)$, that is cubic in sequence length.
Padding to make shapes constant would make things worse.
Thus, naive techniques will not scale.

\textit{With such difficulties in making the training scalable, it may not be surprising that despite the simplicity of \cats, it was not attempted in the community.}
Note that unlike \cats, similar architectures \cite{ho2024block, yu2023megabyte} do not have this problem: computing logits can be naively parallelized due to fixed shapes and self-attention operations scale quadratically due to a single compressed representation for the past.

In \cats, observe that in computing logits chunks $\mbc_i,\mbc_{i+1} \dots \mbc_{\frac{N}{C}}$, one calculates the same key-values for chunk representations $f_\theta(\mbc_j)$ in the decoder, where $j<i$.
This points to repeated and identical computations.
To exploit this observation, we take advantage of a custom attention mask in decoder to calculate logits for all chunks in parallel, and reuse computations done for a past chunk representation to be used for a computations for logits for a future chunk.
To be concrete, once we calculate all chunk representations $f_\theta(\textbf{c}_i)$ in parallel using \texttt{torch.vmap}, we insert $f_\theta(\textbf{c}_i)$s at particular positions in the original sequence: after every chunk $\mbc_i$, we attach its chunk representation.
That is, sequence would look like: $\{\mbc_1,f_\theta(\mbc_1),\mbc_2,f_\theta(\mbc_2),\dots \mbc_i,f_\theta(\mbc_i)\dots\}$.
Now, we pass this sequence into the decoder during training, with a custom attention mask (see Figure \ref{fig:cat_attention_mask}) that allows a token in chunk $\mbc_i$ to attend to previous tokens within that chunk only as well as only to previous chunk representations, which would be $f_\theta(\mbc_{i-1}), f_\theta(\mbc_{i-2})\dots f_\theta(\mbc_1)$ only.
Any token in chunk $\mbc_i$ does not attend to raw tokens outside this chunk.
This implementation allows re-use of key-values for chunk representations $f_\theta(\mbc_i)$ for calculation of logits of future chunks, in parallel, making the training of \cats efficient and scalable.
We utilize the FlexAttention API \cite{dong2024flex} to automatically create a custom kernel for the custom mask (Figure \ref{fig:cat_attention_mask}).
Note that this way of computing logits is quadratic in sequence length but with a constant times better: concretely it is $O(\frac{N}{C}\cdot N + \frac{N}{C}\cdot C^2)=O(\frac{N^2}{C})$, \textbf{which is $C\times$ better than $O(N^2)$} (yellow dots in figure \ref{fig:cat_attention_mask} provides a visual proof for this cost; number of yellow dots are significantly lower than $\frac{N^2}{2}$). Mathematically the cost of attention in \cats decoder is: $\sum_{i=1}^{N}[\frac{i}{C}]+(i \bmod C)+1=O(\frac{N^2}{C})$, where $[.]$ is the floor function, and $\bmod$ is modulo operator.

For a discussion in training throughput, refer to a discussion in Appendix \ref{app:cat_training_throughput}.

\begin{lstlisting}[style=pytorch, caption={Pseudocode for training step}]

def forward(input_ids, targets):

    input_ids = einops.rearrange("b (k c) -> b k c", k=num_chunks, c=chunk_size)

    # calculate f(x)
    # shape of fx: (b, k, D_d)
    fx = torch.vmap(f)(input_ids)

    output_logits = list()
    for i in range(num_chunks): # note that this loop is done in parallel with the custom attention mask presented in the appendix
        # use the previous i+1 fx to predict the current chunk
        # shape of cur_chunk_logits: (b, 1, l, V)
        cur_chunk_logits = phi(input_ids[:, i, :], fx[:, :i+1, :]) 
        output_logits.append(cur_chunk_logits)
    output_logits = torch.cat(output_logits, dim=1) # shape: (b, k, c, V)
    output_logits = einops.rearrange(output_logits, "b k c v -> b (k c) v") # arrange all chunks logits together (or flatten)
    return torch.nn.functional.cross_entropy(output_logits, targets) # return the loss
\end{lstlisting}

\subsection{\cat's training attention mask}
\label{app:cat_attention_mask}
\begin{figure}[h]
    \centering
    \includegraphics[width=0.5\linewidth]{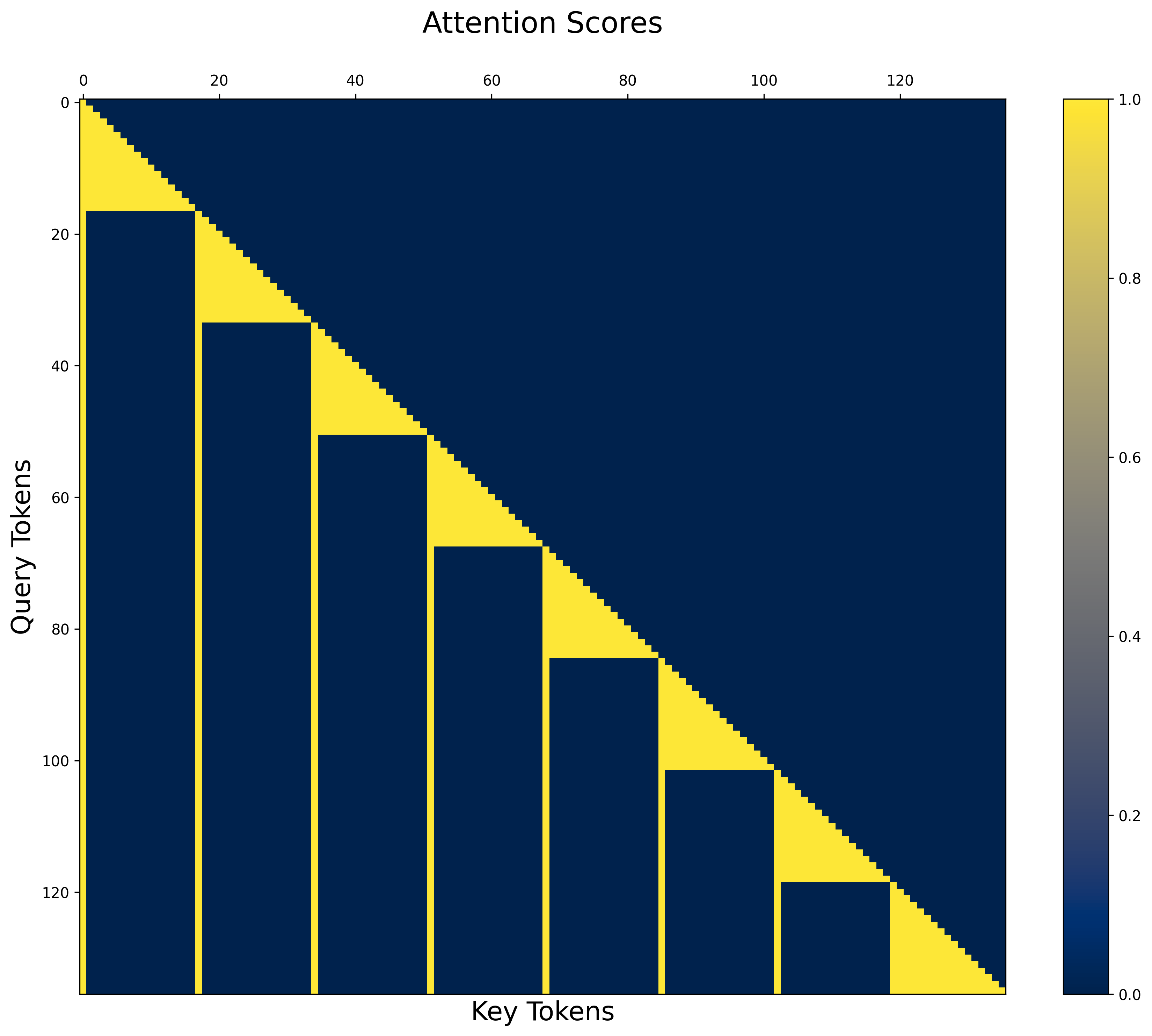}
    \caption{Sequence length is 128, and the chunk size that we use in this particular attention mask is $C=16$.}
    \label{fig:cat_attention_mask}
\end{figure}

Note that attention mask in figure \ref{fig:cat_attention_mask} looks very similar to the attention mask as defined in \cite{child2019generating}, however, in \cat's case: (a) it is not heuristic choice, and (b), tokens in a particular chunk attend to the past $f_\theta(\textbf{c}_i)$ representations obtained by the compressor, rather than the past token embeddings at that position as done in \cite{child2019generating}.

\newpage
\subsection{Generation}

The decoder during generation attends to atmost $\frac{N}{C}+C$ tokens.
Due to compression, \cats can throwaway past chunks of tokens, and only keep their compressed chunk representations in memory.
This straightaway results in a big reduction of memory; the KV cache is slashed by a factor of $C$. 
For even a moderate chunk size of 4, this results in big reductions in memory during generation (Figure \ref{fig:generation_throughput}) compared to a dense transformer.
This slash in memory is accompanied by reduced memory accesses a decoder makes in \cats, which is the major bottleneck during generation.
Costs for self-attention in \cats decoder scale as $O(\frac{N^2}{C})$, which is again, $C\times$ better than $O(N^2)$ for a dense transformer.

Implementing generation is simpler than training and very similar to how it occurs for a dense transformer.
In fact, a pure PyTorch implementation for \cats is on-par with efficient architectures that utilize custom kernels.
We inspire our implementation from: \url{https://github.com/meta-pytorch/gpt-fast}.
Given $i$ chunks of tokens:
firstly, \texttt{torch.vmap} over chunks independently to calculate $f_\theta(\mbc_i)$ in parallel.
Then prefill the decoder's KV cache in parallel with the obtained $f_\theta({\mbc}_i)$s.
Now generate the next chunk ${\mbc}_{i+1}$ autoregressively one token at a time. Note that this uses a simple causal mask since the previous positions are already prefilled with $f_\theta({\mbc}_i)$s, which is required to decode chunk $\mbc_{i+1}$.
    Once all the tokens of the chunk ${\mbc}_{i+1}$ are generated, calculate  $f_\theta({\mbc}_{i+1})$ and prefill the decoder's KV cache just after the position where $f_\theta({\mbc}_{i})$ was cached. 
    Now the KV cache is ready for generation of the next chunk ${\mbc}_{i+2}$ and this process will continue.

This simple implementation enables \cats to be \textbf{$1.4-3.2\times$ faster} than the dense transformer while showcasing \textbf{upto $2.2-9.5\times$ lower total memory usage} as one increases chunk sizes.

\begin{lstlisting}[style=pytorch, caption={Pseudocode for generation}]

# https://github.com/pytorch-labs/gpt-fast/blob/7dd5661e2adf2edd6a1042a2732dcd3a94064ad8/generate.py#L154
def generate_chunk_by_chunk(
    input_ids
):
    # assume input_ids.shape == (batch_size, 1, chunk_size)

    # declare/reset static KV cache, shape: [batch_size, num_chunks + chunk_size, 2, D_d]

    input_pos = 0

    # compress the first chunk (batch_size, 1, chunk_size) -> (batch_size, 1, D_d)
    # get fx for the very first chunk
    fx = f(input_ids) # shape of fx: (batch_size, 1, D_d)
    next_token = prefill(fx, input_pos) # prefill at idx 0 with fx in phi

    new_chunks = list()
    
    for i in range(num_chunks - 1):

        # generate entire chunk using fx that was prefilled earlier in phi
        next_chunk = generate_chunk(next_token)
        new_chunks.append(next_chunk.clone())

        # get new fx
        # compress the new obtained chunk
        fx = f(next_chunk) # (batch_size, 1, chunk_size) -> (batch_size, 1, D_d) 
        
        # prefill again at input_pos
        input_pos += 1
        next_token = prefill(fx, input_pos) # prefill fx at idx `input_pos` in phi
        
    new_chunks = torch.cat(new_chunks)
    return new_chunks

\end{lstlisting}

\subsection{Adaptive \cats training details}
\label{app:adaptive_cat_train_details}

To enable training of adaptive \cats, we made some choices that we now describe.
In every training iteration, we sample a chunk size uniformly at random and perform loss computation.
Further, due to variable size of a chunk in every training iteration, one cannot keep a single projection matrix that projects processed token embeddings in the compressor to a single chunk representation (since shapes for projection matrix would be different for different chunk size).
    One could tackle this by keeping an independent projection matrix for every chunk size, but we found this didn't work well empirically, possibly due to reduced updates for every chunk size's projection weights (only one chunk size's projection weights are updated per iteration; this is not the case with compressor or the decoder, they are updated every iteration).
    Instead, we took inspiration from \cite{beyer2023flexivit} where the authors declared a single projection matrix for all chunk sizes, and then linearly interpolated the matrix to the desired shape depending on the current chunk size. 
    This means the linear interpolation is also under \texttt{torch.autograd} and is optimized so that the final linearly interpolated projection matrix gives a \textit{good} chunk representation for every chunk size.

\subsection{\cat's training throughput analysis}
\label{app:cat_training_throughput}

We make use of FlexAttention API to obtain a custom self-attention kernel specifically for the masking scheme section \ref{fig:cat_attention_mask}.
This fused kernel gives a significant boost in training throughput in self-attention costs compared to using a naive PyTorch masked implementation.

MLPs in a transformer drive the majority of the FLOPs budget during training at smaller sequence lengths \cite{scaling-book}.
At a sequence length of 4096, \cats take $\leq2.35\times$ to train compared to a dense transformer (measured on batch size of $8$ with compressor depth of $3$, decoder depth of $6$, hidden size for compressor $D=1024$ and hidden size for decoder $D_g=2D=2048$ for \cat, compared against dense transformer having depth of $6$ and $D=1024$, on a A100 80 GB PCIe.)
At 16K sequence length, this gap reduces significantly and \cat only costs $\leq1.25\times$ more.

Despite this, \cat amortizes this overhead in two ways: (i) through cheaper inference, which dominates lifetime cost, and (ii) by replacing multiple models with one --- training independent models to cover the same range of inference budgets would require separate pretraining runs for each operating point, costing more than a single \cat.

\subsection{Time taken by each \cat component during generation}
\label{app:time_cat_component}

\blue{
Here we measure time taken by each \cat component during generation, specifically time taken by: decoder attention, decoder FFNs, and time taken by parallel compression. \Cref{tab:time_taken_cat_component} provides these results.
We use the same setup in benchmarking as described in \Cref{sec:experiments}. We use a chunk size of 8 for this ablation.
}

\begin{table}[h]
\centering
\begin{tabular}{lrr}
\toprule
Component & Time (ms) & Percentage (\%) \\
\midrule
Attention in Decoder        & 30,817 & 70.1 \\
FFN in Decoder              & 11,555 & 26.3 \\
Compression in Compressor   & 1,551  & 3.5  \\
\midrule
Total                       & 43,932 & 100.0 \\
\bottomrule
\end{tabular}
\label{tab:time_taken_cat_component}
\end{table}

\newpage
\section{Some ablations on the \cat architecture}
\label{app:cat_ablation}

\subsection{Ablation on hidden size of compressor}
With this ablation, we show that increasing hidden size of the compressor does not help in improving perplexity.
We fix $D_g=1536$ for these experiments.
For this ablation, we use a smaller WikiText-103 dataset.
Both compressor and decoder use the same depth $L=6$.

\begin{table}[H]
\centering
\begin{tabular}{@{}llc@{}}
\toprule
\textbf{Chunk Size $C$} & \textbf{Size of $D_f$} & \textbf{Perplexity} \\
\midrule
\multirow{2}{*}{16}  & $768$   &  17.6 \\
                        & $1536$  & 17.6 \\
\bottomrule
\end{tabular}
\caption{Comparison of choices of hidden size of compressor on WikiText-103 perplexity.}
\end{table}

There is no effect of increasing the hidden size of the compressor. The performance before and after remains the same.

\subsection{Ablation on hidden size of decoder}
\label{app:expressive_decoder}
We ablate on different choices of $D_g$ along with different chunk sizes in \cat\space.
In this setup, we fix $D_f$ in the compressor, and only vary $D_g$ or $C$ (chunk size). 
We use WikiText-103 for these experiments.
In this setup, $D=768$. 
Both compressor and decoder use the same depth of $L=6$.

\begin{table}[H]
\centering
\begin{tabular}{@{}llc@{}}
\toprule
\textbf{Chunk Size $C$} & \textbf{Size of $D_g$} & \textbf{Perplexity} \\
\midrule
\multirow{2}{*}{4}  & $D$   & 19.8 \\
                        & $2D$  & 17.4 \\
\midrule
\multirow{2}{*}{8}  & $D$   & 20.4   \\
                        & $2D$  & 17.7   \\
\midrule
\multirow{2}{*}{16} & $D$   & 20.2   \\
                        & $2D$  & 17.6   \\
\bottomrule
\end{tabular}
\caption{Comparison on choices of chunk sizes and sizes of $D_g$ on WikiText-103 perplexity.}
\end{table}

We observe that we obtain the best perplexities when we $D_g=2D$ for the particular chunk size we are using.
Using this observation, we used this as our \textit{default} configuration for the FineWeb-Edu experiments. 

\begin{table}[H]
\centering
\begin{tabular}{lccccc}
\toprule
\textbf{Model} & \textbf{\shortstack[c]{$D_f$}} & \textbf{\shortstack[c]{$D_g$}} & \textbf{Perplexity} & \textbf{Avg.\ recall} \\
\midrule
Dense & $--$  & $D$  & 21.2 & 23.8 \\
\midrule
\textsc{cat} & $D$  & $D$   & 23.8 & 13.7 \\
\textsc{cat} & $D$  & $2D$  & \textbf{20.7}    & \textbf{19.8} \\
\bottomrule
\end{tabular}
\caption{Impact on perplexity and average recall performance of \textsc{cat} when varying $D_g$. For dense, $D_g$ implies hidden size for itself. Here, $D=1024$.
$D_g=2D$ gives better perplexity and average recall.
We train \cat only at chunk size $C=8$ for these experiments.
All models were trained for 5B tokens with 1K sequence length. Rest of the setup follows Sec. \ref{sec:experiments}.
}
\label{tab:decoder_power}
\end{table}

\subsection{Ablation on depth of the compressor}

We ablate on the depth of the compressor. 
For a fixed chunk-size, $D_f=768$ (compressor embedding size), $D_g=1536$ (decoder hidden size), and a fixed depth of the decoder, we vary the compressor depth. 

\begin{table}[H]
\centering
\begin{tabular}{@{}llc@{}}
\toprule
\textbf{Chunk Size $C$} & \textbf{Depth of Compressor} & \textbf{Perplexity} \\
\midrule
\multirow{2}{*}{8}  & $6$   & 17.4   \\
                        & $3$  & 17.4   \\
\midrule
\multirow{2}{*}{16} & $6$   & 17.8   \\
                        & $3$  & 17.7   \\
\bottomrule
\end{tabular}
\caption{Comparison on choices of depth of the compressor across different chunk sizes $C$ on WikiText-103.}
\end{table}

We have an interesting observation that one can reduce the depth of the compressor without sacrificing on the downstream perplexity.
This could mean one can compress small chunks of tokens without a requiring high capacity.
In our generation benchmarks, we observed that compressor depth play less of a role in latency as compared to the decoder depth (since we compress tokens in parallel using one transformer call).
That being said, compressor depth does play a significant role in training costs (due to the MLP training costs in the compressor).
Therefore, reducing compressor depth goes into overall advantage for the \cat\space architecture.

However, what is the limit, and can one go to even a 1 layer of compressor is an interesting question to ask. 
There might be some lower bound on the compressor depth to start compressing chunks of tokens, but we leave this to future work.

\newpage
\section{More experiment details}
\label{app:training_details}

Here we provide more details about the experiments done in the main text.

\subsection{Baselines}

In this section, we provide details about the models used in our experiments.

\begin{table}[h]
\centering
\begin{tabular}{lccc}
\toprule
\textbf{Model} & \textbf{Total (M)} & \textbf{Embedding (M)} & \textbf{Non-Embedding (M)} \\
\midrule
Dense   & 260         & 50           & 210 \\
Dense-D/2   & 92         & 25           & 70 \\
Mamba2  & 260         & 50           & 210 \\
GDN     & 310         & 50           & 260 \\
GDN-2x     & 310         & 50           & 260 \\

GDN-Hybrid 1:1     & 280         & 50           & 230 \\
GDN-Hybrid 1:1 D/2     & 111         & 25           & 86 \\
GDN-Hybrid 1:4     & 280         & 50           & 230 \\
GDN-Hybrid Global 1:5     & 300         & 50           & 250 \\
GDN-Hybrid Global 1:5 2D   & 820         & 100           & 720 \\
GDN-2x 2D     & 820         & 100           & 720 \\
Sparse-4  & 820         & 100           & 720 \\
Sparse-8  & 820         & 100           & 720 \\
\midrule
\cat-4/8/16/32     & 150 + 820 & 50 + 100   & 100 + 720 \\
\bottomrule
\end{tabular}
\caption{
\blue{Model parameter sizes in millions, separated into embedding and non-embedding parameters.
Parameters for \cats consists of parameters in compressor + parameters in decoder.}
}
\label{tab:model_sizes}
\end{table}

By default, all models below are configured with
$L=12$ layers and $D=1024$ hidden dimension; any deviations are explicitly stated.

\begin{enumerate}
    \item Dense transformer (or Transformer++) \cite{vaswani2017attention, touvron2023llama}: We use rotary position embeddings along with the FlashAttention kernel to perform self-attention. The MLP is a SwiGLU MLP \cite{touvron2023llama}.
    Dense-D/2 uses $2\times$ lower model dimension of $D=512$.

    \item Sparse transformer \cite{child2019generating}: Follows the Dense transformer configuration, except the attention mask used.
    Moreover, we used $D=2\cdot1024=2048$ for this baseline for a fair comparison with \cats.
    We used FlexAttention API to create optimized Flash Attention like kernel for this. We use a stride length of 4 (Sparse-4) and 8 (Sparse-8) that tries to compete with \cat-4 and \cat-8 respectively.

    \item \textsc{mamba2} \cite{dao2024transformers}: The model uses 2 Mamba mixer per layer. 
    All layers use the \textsc{mamba2} block without any mixing any attention. The \texttt{expand} is set to 2, $d_{state}=128$, and convolution $k=4$. 
    Activations used are \texttt{SiLU}. 
    We use the official codebase for \textsc{mamba2} generation throughput and memory benchmarking: \url{https://github.com/state-spaces/mamba} and code from: \url{https://github.com/fla-org/flash-linear-attention} for training.

    \item GatedDeltaNet (\textsc{gdn}) \cite{yang2025gated}: We use the implementation provided at \url{https://github.com/fla-org/flash-linear-attention} for training.
    We use \texttt{head\_dim} as 128 and \texttt{num\_heads} as 8 (same as \textsc{mamba2} above).
    \textsc{gdn-2x} stands for recurrent state size increased by $2\times$.

    \item GatedDeltaNet Hybrids (\textsc{gdn-h}) \cite{yang2025gated}
    We insatiate multiple GatedDeltaNet Hybrids models in our comparison. 
    \textsc{gdn-h 1:1} uses use sliding window layers at every other layer with a sliding window size of $2048$.
    \textsc{gdn-h 1:4} uses the same sliding window layers, but in the ratio of $1:4$ with linear attention.
    \textsc{gdn-h 1:1 D/2} uses the same sliding window layers at every other layer, but the model dimension is scaled down by $2\times$ to $D=512$.

    Finally, \textsc{gdn-h 1:5 G} uses linear-dense attention ratio as 1:5 but with a global attention. \textsc{gdn-h 1:5 2D G} uses linear-dense attention ratio as 1:5 with global attention and uses $2\times$ the hidden size.

\end{enumerate}

\subsection{Datasets}
\label{app:datasets}

Following common practices done in \cite{gu2023mamba, dao2024transformers, arora2024simple, yang2025gated}, we evaluate all models on multiple common sense reasoning benchmarks: PIQA \cite{bisk2020piqa}, HellaSwag \cite{zellers2019hellaswag}, ARC-challenge \cite{arc_challenge}, WinoGrande \cite{sakaguchi2021winogrande} and measure perplexity on WikiText-103 \cite{merity2016pointer}and LAMBADA \cite{paperno2016lambada}.
In Table~\ref{tab:lm_eval}, HS denotes HellaSwag, PQ denotes PIQA, AE denotes ARC-Easy, AC denotes ARC-Challenge, WG denotes Winogrande, OQA denotes OpenBookQA, LMB denotes LAMBADA, Wiki denotes WikiText, and FW denotes FineWeb-Edu.

We evaluate on tasks from LongBench \cite{bai2023longbench} where each abbrevation in table \ref{tab:longbench} stands for: QAS: \texttt{qasper}, MQA: \texttt{multifieldqa\_en}, HQA: \texttt{hotpotqa}, 2WMQ: \texttt{2wikimqa}, TQA: \texttt{triviaqa}, TREC: \texttt{trec} split of LongBench.

To measure real-world recall accuracy, we use datasets used in \cite{arora2024simple, arora2024just}. Namely these consists of SWDE \cite{lockard2019openceres} for structured HTML relation extraction and several question answering datasets including SQuAD \cite{rajpurkar2018know}, TriviQA \cite{joshi2017triviaqa}, DROP \cite{dua2019drop} and FDA \cite{arora2023language}.
Since our pretrained models are small, we use the Cloze Completion Formatting prompts provided by \cite{arora2024just}.

We evaluate on tasks from the needle-in-haystack benchmark RULER \cite{hsieh2024ruler}. 

Additionally, we evaluate on datasets from the LongBench benchmark \cite{bai2023longbench} to evaluate long-context understanding.

\subsection{Generation}
\label{app:generation_details}

Both dense transformer and \cat\space use out-of-the-box FlexAttention API causal dot product kernels in PyTorch.
We use the script provided in \cite{dao2024transformers} to benchmark\footnote{\href{https://github.com/state-spaces/mamba/blob/main/benchmarks/benchmark_generation_mamba_simple.py}{github.com/state-spaces/mamba}} Mamba2 and \texttt{flash-linear-attention} repo to benchmark \textsc{gdn}.
All benchmarks used a prefill of $8$ tokens.
All benchmarks were run using a single NVIDIA A100 80GB PCIe, and use CUDA cache graphs for the next-token prediction.

\textbf{To measure throughput}, we measure the latency at increasing batch sizes (in powers of 2) and compute the maximum tokens per second a model can obtain on a fixed hardware. 
For our experiments, fixed hardware for throughput is a single A100 PCIe 80GB GPU.

\subsection{MQAR setup}
\label{app:sparse_fails_mqar}

We evaluate models on the synthetic multi-associate query recall (MQAR) task, proposed in \cite{arora2023zoology} and further popularized in \cite{arora2024simple}.
All models use depth of 2 layers, and are trained and tested on sequence lengths upto 1024 having the maximum number of key-value pairs possible.
\cat models use a 1 layer compressor, followed by a 2 layer decoder, with a chunk size of 4, both using model dimension of $D=D_d=64$ in this case. 

We use the state size calculations provided in \cite{arora2024simple, arora2023zoology}.

\subsection{Main figure details}
\label{app:figure_details}
\blue{
\Cref{fig:pareto_frontier} reports results at 2K sequence length since both SWDE and FDA datasets have queries with context length upto 2K.
}

%% file: sections/07_extended_related_work.tex
\section{Related Work}
\label{app:extended_related_work}

\begin{table*}[h]
\scriptsize   %
\setlength{\tabcolsep}{3pt}
\renewcommand{\arraystretch}{1.3}
\begin{tabular}{|p{0.18\linewidth}|p{0.10\linewidth}|p{0.09\linewidth}|p{0.09\linewidth}|p{0.14\linewidth}|p{0.09\linewidth}|p{0.10\linewidth}|p{0.10\linewidth}|}
\hline
\textbf{Method} & \textbf{Unrestricted Access to Memory?} & \textbf{Flexible memory?} & \textbf{Scalable training?} & \textbf{Both compute \& memory efficient?} & \textbf{Controllable inference costs?} & \textbf{Mixable with \cat?} & \textbf{Usable in \cat's decoder?} \\
\hline
\textbf{\textit{Dense}}: \cite{vaswani2017attention}
  & \cmark{} %
  & \cmark
  & \cmark
  & \xmark
  & \xmark
  & \cmark
  & \cmark \\
\hline
\textbf{\textit{Sparse Attention}}: \cite{child2019generating}
  & \xmark{} %
  & \cmark
  & \cmark
  & \cmark
  & \xmark
  & \cmark
  & \cmark \\
\hline
\textbf{\textit{NSA}}: \cite{yuan2025native}
  & \cmark{} %
  & \cmark
  & \cmark
  & \xmark
  & \xmark
  & \cmark
  & \cmark \\
\hline
\textbf{\textit{Sliding window Attn.}}: \cite{jiang2023mistral7b}
  & \xmark
  & \xmark
  & \cmark
  & \cmark
  & \xmark
  & \cmark
  & \cmark \\
\hline
\textbf{\textit{Linear Attention}:} \cite{dao2024transformers}
  & \cmark{} %
  & \xmark
  & \cmark
  & \cmark
  & \xmark
  & \cmark
  & \xmark \\
\hline
\textbf{\textit{Recursive compression}}: \cite{chevalier2023adapting}
  & \cmark
  & \cmark
  & \xmark{} %
  & \cmark
  & \xmark
  & \cmark
  & \cmark \\
\hline
\textbf{\textit{MegaByte/Block Transformer}}: \cite{ho2024block, yu2023megabyte}
  & \cmark
  & \xmark
  & \cmark
  & \cmark
  & \xmark
  & \cmark
  & \cmark \\
\hline
\textbf{\textit{\cats}}
  & \cmark %
  & \cmark %
  & \cmark %
  & \cmark
  & \cmark
  & \cmark
  & \cmark \\
\hline
\end{tabular}
\caption{
We categorize existing related work by key properties desirable for an efficient architecture, and indicate whether \cat{} can complement these approaches as a \textbf{meta-sequence mixer}.
\textit{``Both compute and memory efficient?''} signifies savings during inference;
\textit{``Unrestricted Access to Memory''} signifies whether an architecture can freely access any part of the memory in the past, without any artificial restrictions;
\textit{``Mixable with \cat?''} indicates whether \cat{} as a layer be used in these approaches;
\textit{``Usable \cat's compressor/decoder?''} indicates whether the method can serve as a compressor or decoder within \cat. Note that \cat itself can be recursively used as compressor/decoder.
}
\end{table*}

\paragraph{Efficient sequence mixers:} 
Sparse or sliding window attention \citep{child2019generating, zaheer2020big, jiang2023mistral7b} heuristically restrict which tokens are attended to. 
This reduces compute (and sometimes memory), but if the wrong mask is chosen, these methods underperform or require more depth \citep{arora2024simple}. Matching dense transformer quality often requires large windows or composition with dense attention at specific layers \citep{arora2024simple, agarwal2025gpt}.
Linear attention \citep{katharopoulos2020transformers, arora2024simple, dao2024transformers, yang2025gated} replaces softmax with kernelized attention, admitting a recurrent form with constant memory. Recent variants add data-dependent gating \citep{dao2024transformers, yang2025gated}, but all require handcrafted state update rules. The fixed-size recurrent state struggles with long-range recall \citep{arora2024simple, jelassi2024repeat, wen2024rnnstransformersyetkey}, and making these mixers competitive requires careful composition with attention -- a process that involves significant trial-and-error \citep{waleffe2024empirical, qwen_blog_2025, wang2025systematic}.
\cat complements any sequence mixer whose decoding cost depends on sequence length, since instantiating \cat's decoder with such a mixer reduces costs in that mixer due to the compressed sequence.
Consequently, the only strict competitors are sequence mixers with length-independent costs, such as pure linear attention.
Otherwise, any existing efficient sequence mixer can serve in \cat, or be used alongside \cat for test-time flexibility, including hybrid designs (App. \ref{app:cat_as_meta_mixer}).

\textit{Training-free efficiency}: Plethora of works have tackled reducing compute requirements of a transformer in a \textit{post-hoc} manner i.e. after it has been trained using full-attention (also called \textit{training-free} sparse attention) \cite{nawrot2025sparse, li2024snapkv, tang2024quest}.
However, because models are trained dense but run sparse, train-test mismatch can hurt quality. 
Moreover, many target only compute savings and \textit{not} memory savings, which is the main bottleneck for long-contexts.
That being said, these are complementary and can be applied directly to the decoder in \cat, providing additional savings besides what \cat already provides.
We deliberately instantiated \cat with the most-basic and ubiquitous sequence mixer: dense attention for this purpose.
For completeness, we provide a comparison with a training-free method \cite{jiang2024minference} in App. \ref{app:training_free}.

Mixture-of-Experts (MoEs) \citep{shazeer2017outrageously, dai2024deepseekmoe} take a different approach: they increase parameters in feed-forward layers via sparse computation, without increasing inference costs -- making them complementary to any sequence mixer, including those that \cat utilizes in compressor and decoder.

\paragraph{Compressing past context:}
Recurrent compression \citep{Rae2020Compressive, chevalier2023adapting} enables generation of longer sequences on limited compute and memory. However, sequential training is slow and memory-intensive, scaling poorly on modern hardware that favors parallelism. Training recurrent models also poses optimization challenges; \citet{geiping2025scaling} required careful recipes to prevent collapse when scaling up.
Native Sparse Attention (NSA) \citep{yuan2025native} attends to compressed chunks as well as few raw tokens, with compression happening in parallel at every layer. This is similar in spirit to \cat, but NSA retains the full KV cache for the entire context—yielding compute savings but no memory savings during inference.
\textbf{Further, no method here provides test-time control of inference costs, which is the primary goal of \cat.}

\paragraph{Hierarchical transformers:} 
Hourglass architectures \citep{nawrot2021hierarchical, nawrot2022efficient, slagle2024spacebyte} downsample the sequence into coarse tokens, then upsample before decoding. This saves compute during training, but generation still requires memory accesses for all past tokens, especially \textit{fine-grained} ones, which is the main bottleneck.
MegaByte and Block Transformer \citep{ho2024block, yu2023megabyte} model sequences as independent chunks conditioned on a single compressed representation of the past. While this improves efficiency, the fixed-size bottleneck hurts recall even on simple tasks (see App.~\ref{app:block_transformer_fails}). 

Byte Latent Transformer (\textsc{blt}) \cite{pagnoni2025byte} utilizes an \textit{hourglass} model to compress in an \textit{data-dependent} way according to entropy, judged by a separately trained model. 
However, there is no mechanism to change inference costs in \textsc{blt} if the same sequence is fed into the model. 
\cat is actually complementary, where the goal is to provide controllable inference-costs.
In fact, one may instantiate the main network of \textsc{blt} as \cat, and get both data-dependent compression and controllable efficiency in a single model.

In summary, \cat complements most existing (or future) approaches, can extend them, or be mixed with them to unlock test-time control of inference cost.

\paragraph{Adaptive architectures:} \cite{kusupati2022matryoshka, devvrit2023matformer} learns representations during training time that can work at different granularity during test-time, yielding adaptivity to the learned architecture.
However, coarser granularity of \textit{Matryoshka} representations result in loss of language modeling performance (in terms of perplexity) \cite{devvrit2023matformer}.
That being said, one could apply similar approaches to \cats making them complimentary.
\cats use the same high-level approach described in \cite{beyer2023flexivit}: learn a single model that can work for various patch sizes at once depending on the downstream use-case at test-time. However, \cite{beyer2023flexivit} worked with image classification tasks; \cats deal with language modeling and generation.